\documentclass[journal,twoside,web]{ieeecolor}
\usepackage{generic}
\usepackage{cite}
\usepackage{amsmath,amssymb,amsfonts}
\usepackage{graphicx}
\usepackage{algorithm}
\usepackage[hidelinks]{hyperref}
\usepackage{textcomp}
\usepackage{subcaption}
\usepackage{chngcntr}
\usepackage{comment}
\usepackage{apptools}
\newtheorem{theorem}{Theorem}[section]

\newtheorem{proposition}[theorem]{Proposition}

\usepackage{tikz}
\usetikzlibrary{fit,positioning,calc}
\usepackage{algpseudocode}
\def\BibTeX{{\rm B\kern-.05em{\sc i\kern-.025em b}\kern-.08em
		T\kern-.1667em\lower.7ex\hbox{E}\kern-.125emX}}
\newcommand{\linebreakand}{%
  \end{@IEEEauthorhalign}
  \hfill\mbox{}\par
  \mbox{}\hfill\begin{@IEEEauthorhalign}
} 
\makeatother

\def\BibTeX{{\rm B\kern-.05em{\sc i\kern-.025em b}\kern-.08em
T\kern-.1667em\lower.7ex\hbox{E}\kern-.125emX}}

\begin{document}
\title{
\vspace{-0.4cm}
Learning An Active Inference Model of Driver Perception and Control:
Application to Vehicle Car-Following}
\author{Ran Wei,
Alfredo Garcia,
Anthony McDonald,
Gustav Markkula,
Johan Engstrom, Matthew O'Kelly
\vspace{-1cm}
\thanks{This work was supported in part by Army Research Office ARO under grant W911NF-22-1-0213. }
}
\maketitle
\begin{abstract}
In this paper we introduce a general estimation methodology for learning a model of human perception and control in a sensorimotor control task based upon a finite set of demonstrations. 
The model's structure consists of {\em (i)} the agent’s internal representation of how the environment and associated observations
evolve as a result of control actions and {\em (ii)} the agent’s preferences over observable
outcomes. We consider a model's structure specification consistent with {\em active inference}, a theory of human perception and behavior from cognitive science.
According to active inference, the agent acts upon the world so as to minimize {\em surprise} defined as a measure of
the extent to which an agent’s current sensory observations differ from its preferred sensory observations. We propose a bi-level optimization approach to estimation which relies on a structural assumption on prior distributions that parameterize the statistical accuracy of the human agent's model of the environment.
To illustrate the proposed methodology, we present the estimation of a model for
car-following behavior based upon a naturalistic dataset.
Overall, the results indicate that learning active inference models of human perception and control from data is a promising alternative to black-box models of driving.
\end{abstract}

\begin{IEEEkeywords}
Human perception and action, Partially Observable Markov Decision Process, Active inference, inverse reinforcement learning.
\end{IEEEkeywords}

\vspace{-0.4cm}
\section{Introduction}
\label{sec:introduction}
\IEEEPARstart{I}{n} many control tasks requiring mind and motor resources by a human agent, the observation space can be high-dimensional and complex. Empirical evidence indicates that humans agents use a simpler, lower dimensional representation of the environment in sensorimotor control tasks \cite{Badre_2021, DeBeek_2002}. The Bayesian brain hypothesis \cite{Knill_2014, Friston_2012} posits that the human brain uses the information provided by sensory data to update a representation of the world in the form of a conditional probability distribution. To account for the structure of human perception and control in a task involving motor and mind resources, a model must separately describe {\em (i)} the agent’s internal representation of the world as the environment
evolves as a result of control actions and {\em (ii)} the agent’s preferences over observable
outcomes. Equipped with data in the form of demonstrations (i.e. sequences of recorded observation-action pairs), the learning task is to estimate the agent’s preferences as well as its internal representations leading to a behavior policy that best fits data. 

In machine learning, this estimation problem is known as {\em inverse} reinforcement learning (IRL) in an {\em off-line} setting \cite{Osa_2018}. In contrast to reinforcement learning (wherein the goal is to identify a control policy based upon reward and state observations), the goal of IRL is to estimate the reward function and transition probabilities from observed trajectories of state-action pairs\cite{ng2000algorithms, osa2018algorithmic,Abazar_2020}. 
The estimated reward provides an interpretation of the agent's behavior and the reward function can be used to design policies in domains where manual reward specification is difficult, e.g., in autonomous driving \cite{Phan_2023}.
There is a significant literature on identification and estimation of models of human control when the state is observable \cite{Boer_1998,Van_2016, Drop_2018}.
In contrast, model identification and estimation when the state is only partially observable (as in models accounting for human perception) has received less attention.
Notable exceptions include \cite{Baker_2017, Chang_2022, Straub_2022, Pekkanen_2018}. However, the environments considered in these papers are either low dimensional as in \cite{Baker_2017, Chang_2022},  restricted to a linear-quadratic control \cite{Straub_2022} or customized for a specific control task \cite{Pekkanen_2018}.

In this paper, we introduce a Bayesian estimation methodology for learning a structural model of perception and control in general control tasks in higher dimensional settings. We use the formalism of a Partially Observable Markov Decision Process (POMDP) in which the agent's preferences are modeled by a reward function and the agent's internal representation of the environment consists of observation and transition probabilities.
However, a POMDP model of a human agent's perception and control policy based solely on demonstrations is in general {\em non-identifiable}, i.e. there may be several different combinations of reward and internal model of the environment that rationalize the same demonstrations dataset. This is because in planning control tasks, different combinations of reward and internal dynamics model could result in the same inter-temporal reward trade-offs.
To address this issue, we make a structural assumption on prior distributions that parameterize the statistical accuracy of the human agent's model of the environment.
Specifically, we assume
({\em i}) the agent's preferences and model of the environment are independent and ({\em ii}) the distribution of the model of the environment parameters concentrates on values with higher fit to the data (i.e. higher log-likelihood).
In words, this assumption restricts our estimation to agents with reasonably accurate models of the environment whose preferences over states of the world are not determined by their perception of the environment. This allows us to formulate the {\em Maximum A Posteriori} (MAP) estimator as the solution to a bi-level optimization problem. The upper-level problem is the maximization of the posterior distribution and the lower-level problem is the computation of optimal policy for the given reward {\em and} model of the environment. We approximate the solution to this bi-level optimization problem by a stochastic gradient algorithm with a nested policy optimization step.

To illustrate the proposed methodology, we consider an application to
highway driving  using a naturalistic dataset.
We specify the structure of the model in accordance to ``active inference" \cite{friston2010free, Parr_2022, Maisto},
a novel framework for modeling human perception and behavior in sensorimotor control tasks \cite{Engstrom_2024}. Active inference is related to the Bayesian Brain hypothesis that posits prediction as the fundamental task of cognition \cite{Clark_2015, Hohwy_2015}, i.e. the brain minimizes prediction error by updating beliefs about the states of the world consistent with data. Active inference takes a conceptual leap from this view in that minimizing prediction error can also be attained by both updating beliefs and acting upon the world to approximately induce a preferred distribution of the states of the world. The active inference framework is summarized by a principle of {\em free energy minimization}: forward (action) and backward (belief) updating processes work in tandem to minimize ``surprise" with respect to a preferred belief distribution about the states of the world.

Our ultimate goal is to provide a model that is {\em interpretable} --in terms of a cognitive model of perception and action-- and that exhibits a statistical performance that is similar or arguably superior to other black-box models. To this end, 
we compare the learned structural model of perception and action (with an active inference reward specification) with two baseline models based on Behavior Cloning (BC), a common machine learning approach to driving behavior modeling \cite{suo2021trafficsim, igl2022symphony, codevilla2019exploring, zhou2017recurrent}, and the Intelligent Driver Model (IDM), a class of rule-based models widely used by traffic simulation software \cite{treiber2000congested, treiber2013microscopic, kesting2009agents}. 
It is important to emphasize that the naturalistic dataset does not include collisions and only includes relatively few extreme observations (e.g. extremely low relative distance or high relative velocity). Thus, our model only provides an account of driver perception and control behavior in {\em average} conditions and our testing of model performance focuses on {\em aggregate} measures.
The results indicate that the active inference-based model outperforms those obtained by imitation learning-based models from the machine learning literature. However, the learned model is inaccurate in extreme scenarios that are poorly covered in the dataset and does exhibit higher collision rates than IQM when tested online. This is due to the limitations of the dataset which poorly covers extreme scenarios so that the learned model does poorly in extreme scenarios due to distribution shift \cite{Quionero_2009, levine2020offline}.
Overall, the results indicate that learning active inference models from data is a promising alternative to black-box models of driving as it provides a way to trace driving behaviors back to a human drivers' perception and preferences.

The structure of this paper is as follows. In section II, we start by describing a Partially Observable Markov Decision Process (POMDP) of perception and control. In section III, we describe the inverse estimation problem, i.e. based upon sequences of observations and implemented actions to estimate the primitives of the POMDP model (reward, partially observable state transition and observation probabilities). In section IV, we describe the specification of reward function based upon active inference, a novel framework for cognition and behavior.
In section V, we describe the application of the proposed estimation algorithm to obtain an active inference model  for car-following behavior by human drivers. We compare the active inference model with Behavior Cloning (BC) and the Intelligent Driver Model (IDM). Finally, in section VI, we close with concluding remarks about the promise and challenges of learning perception \& control models based upon naturalistic datasets. 

\section{A POMDP Model of Perception and Control}
\label{Sec:Model}

We start by providing a description of a partially Observable Markov Decision Process (POMDP) of perception and control. This encompasses neuroscience modeling frameworks for human perception and action in sensorimotor tasks such as {\em active inference} \cite{friston2010free} and the {\em expected value of control} (EVC) \cite{Shenhav_2017}. 

In the proposed POMDP framework (see Figure \ref{graphical}), the human agent maintains an {internal} model of the world (or representation) so that high-dimensional observations $o_t \in O\subset \mathbb{R}^n$ (sensory stimuli) are represented with a lower dimensional {\em hidden} state $s_t \in S \subset \mathbb{R}^m$ where $S$ is the state space and $m<<n$. 
If the hidden state is $s_t$ and $a_t \in A$ is implemented, the agent accrues a reward $r(s_t, a_t)$. The agent's internal representation includes state dynamics, i.e. a transition to a new state $s_{t+1}$ takes place with probability $\mathbb{T}(s_{t+1}|s_t,a_t)$ and a new observation $o_{t+1}$ is obtained with probability $\mathbb{O}(o_{t+1}|s_{t+1})$. 

After $t>0$ time periods, the observable history of observations and actions is denoted by 
$$h_t:=\{o_t,...,o_0, a_{t-1},...,a_0\}\in H_t \subset O^{t+1}\times A^t.$$ 
We consider randomized or stochastic policies $\pi$ that are adapted to the history of the process, i.e. given history $h_t$ action $a \in A$ is implemented with probability $\pi(a|h_t) \in [0,1], a \in A$ and $\sum_{a \in A} \pi(a|h_t)=1$ for all $h_t \in H_t$.

In the proposed POMDP model of perception and action, the human agent aims to maximize the expected value of discounted reward net of information processing costs:
\begin{equation}    
U_{\tau}(h_{\tau})\triangleq \sup_{\pi \in \Pi} \mathbb{E}\Big[ \sum_{t\geq \tau} \gamma^{t-\tau} [r(s_{t}, a_{t})-c(\pi(\cdot|h_t))]\Big]
\label{U}
\end{equation}
where $\gamma \in (0,1)$ is the discount factor, $\Pi$ is the set of randomized policies that are adapted to the history process and $c(\pi(\cdot|h_t))$ is a per-period {\em information processing cost}.
As human agents may differ in their ability to process task-relevant information or to attend to the task at hand \cite{tishby2011information, Ortega_2013,Matejka_2015,Fudenberg_2015, Hansen_2018}, the cost $c(\pi(\cdot|h_t))$ models the fact that {\em low} entropy behavioral policies are consistent with {\em high} information processing effort or attention.

The combination of additive reward structure and Markovian dynamics allows for a recursive characterization of the optimal policy as follows:
\begin{align}
 U_{t}(h_t) &= \max_{\pi(\cdot|h_t)} \Bigg \{ \sum_{a_t}\sum_{s_t} r(s_t, a_t)b_{t}(s_t)\pi(a_t|h_t) -c(\pi(\cdot|h_t)) \nonumber \\
 &+ \gamma \sum_{o_{t+1}}\sum_{a_t} \mathbb{P}(o_{t+1}|h_t, a_t)\pi(a_t|h_t)  U_{t+1}(h_{t+1}) \Bigg \}. \label{model}  &
\end{align}

Let $b_{t}\in \Delta^S$ denote the Bayes updated belief distribution on the state, i.e. $b_{t}(s):=\mathbb{P}(s_t=s|h_t)$, where $\Delta^S$ is the set of probability distributions on state space $S$.

\begin{figure}
\centering 
  \begin{tikzpicture} 
    [node distance=12.5 pt and 24pt, every node/.style={draw, circle, inner sep=0pt}, minimum size=8mm]

    \tikzstyle{obs}=[circle, draw, minimum size=8mm, thick, draw=black!80, fill=black!20, node distance=3mm] 
    \tikzstyle{label}=[rectangle, draw, minimum size=7mm, thick, draw=black!0, node distance=0mm] 

    \node [] (st) {\normalsize $s_t$};
    \node (s0) [left=0.75cm of st] {\normalsize $s_{t-1}$};
    \node [right=.75cm of st] (s1) {\normalsize $ s_{t+1}$};
    
    \node [obs, above right=.5cm and .3cm of st] (ot) {\normalsize $o_t$};
    \node [obs, above right=.5cm and .3cm of s0] (o0) {\normalsize $o_{t-1}$};
    \node [obs, above right=.5cm and .3cm of s1] (o1) {\normalsize $ o_{t+1}$}; 

    \node [below left=.3cm of s0] (b) {\normalsize $b_{t-1}$};  
    \node [below right=.3cm of s0] (b0) {\normalsize $b_{t}$};  
    \node [below right=.3cm of st] (bt) {\normalsize $b_{t+1}$};
    \node [below right=.3cm of s1] (b1) {\normalsize $b_{t+2}$};    
    
    \node [obs, below=1.15cm of s0] (a0) {\normalsize $a_{t-1}$};  
    \node [obs, below=1.15cm of st] (at) {\normalsize $ a_t$};  
    \node [obs, below=1.15cm of s1] (a1) {\normalsize $a_{t+1}$};
    \node[label,right = 0.5 cm of s1] (s2) {\footnotesize $\cdots$};
    \node[label,left = 0.35 cm of b] (b2) {\footnotesize $\cdots$};
    
    \path [->]
    (s0) edge (st)
    (b)  edge (b0)
    (b)  edge (a0)
    (st) edge (s1)
    (s0) edge (o0)
    (st) edge (ot)
    (s1) edge (o1)
    (o0) edge (b0)
    (ot) edge (bt)
    (o1) edge (b1)
    (b0) edge (bt)
    (bt) edge (b1)
    (b0) edge (at)
    (bt) edge (a1)
    (s1) edge (s2)
    (b2) edge (b);

     \path [->,draw] (at) edge[bend right] (s1);
    \path [->,draw] (a0) edge[bend right] (st);
    
    \draw[thick, dotted, rounded corners=15pt] ($(o0.north west)+(-3.5,.4)$)  rectangle ($(o1.south east)+(.3,-0.2)$);
    \node[label, left=of o0] [left = 1.15 cm of o0]{\footnotesize Observations};

    \draw[thick, dotted, rounded corners=15pt] ($(s0.north west)+(-2.6,0.2)$)  rectangle ($(b1.south east)+(0.6,-0.2)$);
    \node[label] [left= 0.10 cm of s0]{\footnotesize Internal Model};
    
    \draw[thick, dotted, rounded corners=15pt] ($(a0.north west)+(-2.6,0.3)$)  rectangle ($(b1.south east)+(0.4,-1.4)$);
    \node[label, left=of a0] [left = 1.15 cm of a0] {\footnotesize Action};
    \end{tikzpicture}%
\vspace{-0.10cm}
\caption{{\small Graphical Model of Perception and Control.}} 
\label{graphical}
\vspace{-0.7cm}
\end{figure}

Let us denote by $\sigma(o_{t+1}|b_{t},a_{t})$ the probability of recording observation $o_{t+1}$ when action $a_t$ is implemented and the current belief distribution is $b_t$, i.e.
\begin{align*}
\sigma(o_{t+1}|b_{t},a_{t}) &:=
\sum_{s_{t+1}}\sum_{s_t}\mathbb{O}(o_{t+1}|s_{t+1}) \mathbb{T}(s_{t+1}|s_t,a_t)b_{t}(s_t)
\end{align*}
Using standard POMDP arguments in Proposition 1 below we show that with no loss of optimality, the search for optimal policy can be restricted to {\em Markovian} policies, say $\Pi^M \subset \Pi$ that only depend $b_{t}$, the current Bayes updated belief as opposed to the whole history $h_t\in H_t$. 

\begin{proposition}
Let $V_{t}(b)$ be recursively defined as follows:
\begin{align*}
    &V_{t}(b) = \max_{\pi(\cdot|b)} \Big\{\sum_s\sum_{a} r(s, a)\pi(a|b)b(s) -c(\pi(\cdot|b)) \nonumber \\
    &~~~~+ \gamma\sum_{a}\sum_{o'}\sigma(o'|b,a)\pi(a|b) V_{t+1}(b') \Big \}
\end{align*}
where $b'(s) =\mathbb{P}(s_{t+1}=s|h_t\cup(a,o'))$, i.e. the resulting Bayes update after action $a$ is implemented and observation $o'$ are recorded. Then, the Bayes updated belief $b_t=\mathbb{P}(\cdot|h_t)$ is a sufficient statistic for solving \eqref{model}, i.e. $U_{t}(h_t)=V_{t}(b_t)$ for all $h_t$.
\label{SufficientTheorem}
\end{proposition}
\begin{proof} See Appendix.
\end{proof}

We now state and prove the {\em soft} Bellman equation for the value function $V_t(b)$ when the information processing cost is proportional to the Kullback-Leibler divergence between the control policy and a default policy $\pi^{0}$, i.e. $c(\pi(\cdot|b_t)) =  
\alpha \mathcal{D}_{KL}(\pi(\cdot|b_t)||\pi^0(\cdot|b_t))$ where:
\begin{align}
 \mathcal{D}_{KL}(\pi(\cdot|b)||\pi^0(\cdot|b)) &:= \sum_{a \in A} \pi(a|b)\log \frac{\pi(a|b)}{\pi^0(a|b)}.
 \label{KL_divergence}
 \end{align} 
 and $\alpha>0$.

Let $\mathcal{Q}$ be the Banach space of bounded, measurable functions $Q: \Delta^S \rightarrow \mathbb{R}$ under the supremum norm $||.||$. Define the {\em soft} Bellman operator $\mathcal{B}: \mathcal{Q} \rightarrow \mathcal{Q}$ by 
\begin{align}
    [\mathcal{B}Q](b,a):= \sum_s r(s,a)b(s) + ~~ \nonumber \\
    \gamma \sum_{o'}\sigma(o'|b, a)\alpha \log \Big(\sum_{a'}\pi^0(a'|b')\exp\big( \frac{1}{\alpha}Q(b',a')\big)\Big), 
    \label{model4}
\end{align}
where $b'$ is the resulting Bayes update after action $a$ and observation $o'$ are recorded. 
\begin{theorem}
(a) $\mathcal{B}: \mathcal{Q}\rightarrow \mathcal{Q}$ is a contraction mapping with modulus $\gamma \in (0,1)$ with unique fixed point $Q^*$, i.e. 
\begin{align*}
    Q^{*}(b,a)&= \sum_s r(s,a)b(s) + \nonumber \\
    & \gamma \sum_{o'}\sigma(o'|b, a)\alpha \log \Big(\sum_{a'}\pi^0(a'|b')\exp\big( \frac{1}{\alpha}Q^{*}(b',a')\big)\Big),
\end{align*}
(b)
\begin{align*}
V^*(b)  &=\max_{\hat{\pi}(\cdot|b)}\Big[\sum_a \hat{\pi}(a|b)Q^*(b,a)-\alpha \mathcal{D}_{KL}\big(\hat{\pi}(\cdot|b)||\pi^0(\cdot|b)\big)\Big] \\
&=\alpha \log \sum_a \pi^0(a|b)\exp\big(\frac{1}{\alpha} Q^*(b,a)\big)
\end{align*}
(c) the optimal policy is of the form:
\begin{align}
  \pi^*(a|b)=\frac{\pi^0(a|b)\exp\big({\frac{1}{\alpha}Q^*(b,a)}\big)}{\sum_{a' \in A}\pi^0(a'|b)\exp\big({\frac{1}{\alpha}Q^*(b,a')}\big)}.  
\end{align}
\label{CCPTheorem}
\end{theorem}
\begin{proof} See Appendix. \end{proof}

\textbf{Remark 1}: Note that as $\alpha \rightarrow +\infty$, information processing effort is arbitrarily costly and in the limit, the agent implements the default policy $\pi^* \rightarrow \pi^0$. Conversely, as $\alpha \rightarrow 0^+$, we recover the optimal solution without information processing cost since $V^*(b) \rightarrow \max_{a\in A} Q^*(b,a)$.

\textbf{Remark 2}: In the remainder of the paper we shall use $\alpha=1$ and the default policy is the uniformly random policy $\pi^0(a|b)=\frac{1}{|A|}$. With these choices the optimal policy takes the form:
\begin{align}
  \pi^*(a|b)=\frac{\exp Q^*(b,a)}{\sum_{a' \in A}\exp Q^*(b,a')}.  
\end{align}

\textbf{Remark 3 (Finite Horizon)}: It can be easily verified that Proposition 1 and Theorem 1 continue to hold for the case in which the controller is solving a finite horizon problem. Evidently, the results in this case require that the state-action function $Q_{t}$ and the conditional choice probabilities $\pi_{t}$ are time-dependent $t$. Formally, for a planning horizon of length $H>0$, the optimal policy at time $t\in\{0,1,\dots, H\}$ is of the form:
\begin{align}
\pi_{t,H}^{*}(a|b)&=\frac{\exp Q_{t,H}^{*}(b,a)}{\sum_{a' \in A}\exp Q_{t,H}^{*}(b,a')}
\label{receding_soft_policy}
\end{align}
and
\begin{align}
Q_{t,H}^*(b,a)&=\sum_s r(s,a)b(s) + \sum_{o'}
\sigma(o'|b, a)V^*_{t+1,H}(b')
\label{soft_optimal_value} 
\end{align}
where $b'$ is the resulting Bayes update after action $a$ and observation $o'$ are recorded and
\begin{align}
V^*_{t+1,H}(b') & =  \log \Big(\sum_{a'}\exp Q^*_{t+1,H}(b',a')\Big)~~~~t\leq H-1 \label{finite_horizon}
\end{align}
and $
V^*_{H+1,H}=0$.

\section{Estimation Methodology}
Equipped with a model of perception and action as described in the previous section, we consider the estimation of the primitives based upon demonstrations, that is, sequences of observations and implemented actions of the form $\tau = \{(o_0,a_0),(o_2,a_1),\dots,(o_{T},a_T)\}$. We shall denote by $\mathcal{D}$ the finite dataset of distinct sequences of observation-action pairs. 

The primitives of the perception \& control model are parametrized as follows:

\begin{itemize}
    \item 
{\em Perception}: We assume the agent's internal representation of hidden state dynamics and observation probabilities is parametrized with $\theta_1 \in \mathbb{R}^p_1$ so that the likelihood of observation $o_{t+1}$ given beliefs $b_t$ and action $a_t$ is $ \sigma_{\theta_1}(o_{t+1}|b_{t},a_t)$.

\item {\em Preferences}: A reward function $r_{\theta_2}(b,a)$ which is parametrized by $\theta_2 \in \mathbb{R}^p_2$. 
\end{itemize}

Assuming the data is generated by an agent who uses a {\em receding horizon} plan with horizon $H$ according to \eqref{receding_soft_policy}, the log-likelihood of a sequence $\tau \in \mathcal{D}$ can be written as
\begin{align*}
\mathbb{P}(\tau|\theta)&=  \prod_{t=0}^T\Big(\pi^{*}_{\theta}(a_{t}|b_{\theta_1,t})\mathbb{P}\big(o_{t+1}|h_{t} \cup 
\{a_t\}\big)\Big)
\end{align*}
where to alleviate notation we write $\pi^{*}_{\theta}(\cdot|b)$ to refer to the first-period optimal policy with a planning horizon $H>0$. $\mathbb{P}(o_{t+1}|h_{t} \cup 
\{a_t\})$ is the external observation-generating distribution that is \emph{independent} of the agent's internal representation.
Hence, the log-likelihood of dataset $\mathcal{D}$ can be written as:
{\small
\begin{align}
\log \mathbb{P}(\mathcal{D}|\theta)&= \log \prod_{\tau \in \mathcal{D}} \mathbb{P}(\tau|\theta) \nonumber \\
&=\mathbb{E}_{\tau \sim \mathcal{D}}\Big[ \sum_{t=0}^T\log\Big(\pi^{*}_{\theta}(a_{t}|b_{\theta_1,t})\mathbb{P}\big(o_{t+1}|h_t \cup \{a_t\}\big)\Big)\Big]|\mathcal{D}| \\
&=\mathbb{E}_{\tau \sim \mathcal{D}}\Big[ \sum_{t=0}^T\log\pi^{*}_{\theta}(a_{t}|b_{\theta_1,t})\Big]|\mathcal{D}| + \mbox{constant} \label{likelihood}
\end{align}
}
where the expectation is taken with respect to the empirical measure $\bar{\mathbb{P}}(\tau)=\frac{1} {|\mathcal{D}|}$ and
\begin{align}
\pi_{\theta}^*(a|b)& =\frac{\exp{Q_{\theta}^{*}(b,a)}}{{\sum_{a' \in A} \exp {Q_{\theta}^{*}(b,a')}}} \label{softmax}
\end{align}
Condition \eqref{softmax} imposes model in the form of the first period policy of a receding horizon plan. 

We take a Bayesian approach to finding an estimator and make an additional assumption on the structure of the prior distribution of parameters denoted by $P(\theta)$: 

{\bf Assumption 1}:
(a) $P(\theta) = P(\theta_1)P(\theta_2)$. (b) The distribution of $\theta_1$ is of the form:
\begin{align}\label{eq:aif_prior}
    P(\theta_1) \propto \exp \Big(\lambda  \mathbb{E}_{\tau \sim \mathcal{D}}\big[\sum_{t=0}^{T} \log\sigma_{\theta_1}(o_{t+1}|b_{\theta_1,t}, a_{t})\big]|\mathcal{D}|\Big)
\end{align}
for some $\lambda>0$.

Assumption 1(a) restricts our estimation to agents whose preferences (parameterized by $\theta_2$) over states of the world are not determined by their perception of the environment (parameterized by $\theta_1$).
Under assumption 1(b) on the prior distribution, parameter values $\theta_1$ with higher fit to the sequences of observations in the data are more likely. Increasing values of $\lambda$ imply the agent has (a priori) an increasingly accurate model of the environment.

Assuming a uniform prior $P(\theta_2)$ on a compact subset $\Theta_2 \subset \mathbb{R}^p_2$, the log of the posterior distribution can be written as:

{\small
\begin{align}
\log P(\theta|\mathcal{D}) &= \log P(\mathcal{D}|\theta) + \log P(\theta_1) +\mbox{constant} \nonumber \\
&=
 \mathbb{E}_{\mathcal{D}}\Big[\log \sum_{t=0}^T\pi^{*}_{\theta}(a_{t}|b_{\theta_1,t})\Big]|\mathcal{D}| + \nonumber \\
 &~~\lambda \mathbb{E}_{\mathcal{D}}\Big[ \sum_{t=0}^{T}\log \sigma_{\theta_1}(o_{t+1}|b_{\theta_1,t}, a_{t})\Big]|\mathcal{D}| + \mbox{constant}
\label{log_likelihood}
\end{align}
}

We are ready to formulate the estimation problem as the following bi-level optimization problem:
\begin{align}
\max_{(\theta_1,\theta_2)} & ~~\mathbb{E}_{\mathcal{D}}\Big[\log \sum_{t=0}^T\pi^{*}_{\theta}(a_{t}|b_{\theta_1,t}) + \lambda\sum_{t=0}^{T}\log \sigma_{\theta_1}(o_{t+1}|b_{\theta_1,t}, a_{t})\Big] \label{formulation}\\
\mbox{s.t.} &~~~~ \pi^{*}_{\theta} = \arg \max_{\pi \in \Pi^H} \mathbb{E}\Big[\sum_{h\leq H}[r_{\theta}(b_h, a_h) - \log \pi(\cdot|b_h)]\Big]
\nonumber
\end{align}
where here again we write $\pi^{*}_{\theta}(\cdot|b)$ to refer to the first-period optimal policy with a planning horizon $H>0$ with initial belief $b$. The algorithm for approximating a solution, say $\widehat{\theta}$, to \eqref{formulation} is described in Algorithm 1 below. The estimated model structure is summarized as follows:
\begin{center}
\begin{tabular}{ |p{3cm}||p{5cm}| }
 \hline
 \multicolumn{2}{|c|}{Structural Model of Perception and Control} \\
 \hline
Perception &  \\
 \hline
 Observations   & $\mathbb{O}_{\widehat{\theta}_1}(o_t|s_t)$    \\
 Transitions & $\mathbb{T}_{\widehat{\theta}_1}(s_{t+1}|s_t,a_t)$       \\ 
 Generative Model &
$\sigma_{\widehat{\theta}_1}(o_{t+1}|b_{t},a_{t}) 
$
 \\
 \hline
 Control & \\ \hline
 Preferences (reward) & $r_{\widehat{\theta}_2}(s_t,a_t)$ \\
 Control Policy   &  $\pi_{\widehat{\theta}}(a_t|b_t) =\frac{\exp Q^*_{\widehat{\theta}}(b_t,a_t)}{\sum_{a' \in A}\exp Q^*_{\widehat{\theta}}(b_t,a')}$ \\
 \hline
\end{tabular}
\end{center}

\begin{algorithm}[!htb]
\caption{Bayesian MAP Estimation of Perception \& Control Model}\label{algo_btom}
\begin{algorithmic}[1]
\Require Dataset $\mathcal{D} = \{\tau\}$, perception model $\sigma_{\theta_1}(o'|b, a)$, preference model $r_{\theta_2}(b, a)$, initial value $\theta_0=(\theta_{1,0},\theta_{2,0})$, hyperparameter $\lambda>0$ and learning rate $\rho>0$.
\For{$k=0:K$}
    \State Compute the optimal policy $\pi^*_{\theta_{k}}$ using value-iteration
    \State Evaluate the log posterior $\log P(\theta_k|\mathcal{D})$ according to \eqref{log_likelihood}.
    \State Compute the gradient of $\nabla_{\theta} \log P(\theta_k|\mathcal{D})$
    \State Perform parameter update $$ \theta_{k+1}=\theta_{k}+\rho\nabla_{\theta} \log P(\theta_{k}|\mathcal{D})$$
    \vspace{-0.5cm}
\EndFor
\end{algorithmic}
\end{algorithm}

\section{An Active Inference Specification}

In this section we describe a specification of the reward function consistent with active inference \cite{friston2010free}. 
Active inference is a novel framework for cognition and behavior according to which the agent jointly {\em perceives} and {\em acts} upon the world so as to maximize the match between {\em perceived} vs {\em preferred} states of the world.

The process of matching the {\em perceived} vs {\em preferred} distribution of the states of the world follows a principle of {\em free energy minimization}: forward (action) and backward (belief) updating processes work in tandem to minimize a measure of ``surprise" or free energy. For backward (belief) updating, free energy is minimized when the agent's belief distribution $b_t$ corresponds to the Bayes updated belief distribution on the state $s_t$.
For forward (action) selection processes, surprise is measured with respect to a {\em preferred} distribution $\Tilde{P}(s_{t+1})$ over states of the environment. 
The immediate ``surprise" associated with action $a_t$ when current beliefs are $b_t$ is quantified by the {\em expected free energy} defined as:
\begin{align}
\vspace{-.1cm}
    EFE(b_t,a_t) & = 
\mathbb{E}\big[D_{KL}\big(b_{t+1}||\Tilde{P}\big)\big] + \mathbb{E}\big[\mathcal{H}(\mathbb{O}(\cdot|s_{t+1}))\big]
    \label{free_energy}
\end{align}
where the expectation is taken with respect to $o_{t+1}\sim\mathbb{O}(\cdot|s_{t+1}),s_{t+1}\sim\sum_s\mathbb{T}(\cdot|s,a_t)b_t(s)$ with $$b_{t+1}(s)= \mathbb{P}(s_{t+1}=s|h_t \cup \{a_t,o_{t+1}\})$$ and $\mathcal{H}(\mathbb{O}(\cdot|s_{t+1}))$ is the entropy of the resulting generative model of observations, i.e.:
$$
\mathcal{H}(\mathbb{O}(\cdot|s_{t+1})):=-\sum_{o'}\mathbb{O}(o'|s_{t+1})\log \Big(\mathbb{O}(o'|s_{t+1})\Big).
$$
The first term in (\ref{free_energy}) quantifies the extent to which the belief distribution on the states of the world $b_{t+1}$ (resulting from implementing action $a_t$ and recording observation $o_{t+1}$) differs from the preferred distribution of the states of the world $\Tilde{P}(\cdot)$. This term is usually referred to as ``risk" because of its relationship to the deviation from an agent's goal \cite{tschantz2020learning}. Selecting policies that generate preferred observations minimizes risk. 
The second term in (\ref{free_energy}) is a measure of the observation uncertainty induced by action $a_t$.
This term is referred to as ``ambiguity" and represents the value of obtaining reliable information that may help to resolve uncertainty about future states \cite{Parr_2022, tschantz2020learning}. Defining ambiguity hinges on having a model of the world.

In \cite{Shin_2022} an interpretation of active inference (when the state is observable) is given in terms of Markov decision processes. In a similar manner, 
by setting the reward function as $r(b_t,a_t):=-EFE(b_t,a_t)$, the active inference model can be seen as a particular instance of the class of POMDP models described in section \ref{Sec:Model} \cite{wei2024value}. However, the ability to consider trade-offs between the described measures of risk vs. ambiguity presents an advantage of the active inference formulation compared to traditional RL/IRL formulations.

\section{Application: Learning a Model of Perception and Control in Car Following Behavior}
In this section, we describe the application of Algorithm 1 to estimate an active inference model  for car-following behavior by human drivers. \footnote{The source code is available at \url{https://github.com/ran-weii/interactive_inference}.}
Computational models of human performance in such task have been amply studied by traffic engineers and psychologists, see e.g., \cite{Meirav_2001,Hamdar_2008, Hamdar_2015, Siebert_2017}.
However, our goal here is to {\em learn} a model that is motivated by cognitive science (active inference) based upon a naturalistic dataset of task demonstrations. In this sense, the closest paper to our work is \cite{Pekkanen_2018} which assumes the agent's decisions are based upon a {\em state} estimate (speed, relative speed and distance) and a predictive model of the lead vehicle. In contrast, in the proposed POMDP model, the variables speed, relative speed and distance are {\em observations} which are used by the agent to form {\em current} and {\em future} beliefs about the states of the environment which are discrete.\footnote{In this sense, the generative model in the proposed POMDP model can be seen as categorical. There is evidence to support the categorical nature of human perception in psycho-physical experiments \cite{Reed_1972} that are simpler than the one considered in this paper}. In addition, the policy in \cite{Pekkanen_2018} is deterministic and the model is not based on agent's preferences.

We compare the active inference model, referred to as Active Inference Driving Agent (AIDA), with two baseline models: Behavior Cloning (BC), a common machine learning approach to driving behavior modeling \cite{suo2021trafficsim, igl2022symphony, codevilla2019exploring, zhou2017recurrent}, and the Intelligent Driver Model (IDM), a rule-based model used by most traffic simulation software \cite{treiber2000congested, treiber2013microscopic, kesting2009agents}. We begin by describing the baseline models and the dataset used to estimate the parameters of the models. We then describe the protocols for evaluating the models' ability to replicate human driving behavior in the dataset. Lastly, we present the model evaluation results and demonstrate AIDA's interpretability advantages. Implementation details are provided in Appendix \ref{appx:implementation}.

\vspace{-0.75cm}
\subsection{Models and parameterization}
\textbf{Behavior Cloning:} BC trains neural networks to map observations or a history of observations to control actions in the dataset. The policy parameters, denoted with $\theta$, are estimated using maximum likelihood estimation of the dataset actions:
\begin{align}\label{eq:bc_objective}
    \max_{\theta} \mathcal{L}(\theta) = \mathbb{E}_{ \mathcal{D}}\left[\sum_{t=0}^{T} \log \pi_{\theta}(a_t|h_t)\right]
\end{align}
We implement two BC approaches in this work, a standard multi-layer neural network approach (BC-MLP) and a recurrent neural network approach (BC-RNN). These approaches are strong baselines for simulated driving agents.
\cite{codevilla2019exploring, zhou2017recurrent, bhattacharyya2020modeling, kuefler2017imitating}.

\textbf{Intelligent Driver Model:} The IDM \cite{treiber2000congested} accepts observations of vehicle speed $v$, relative speed to the lead vehicle $\Delta v$, and distance headway to the lead vehicle $d$ as inputs and outputs acceleration $a$ (the control action) using the following rule:
\begin{align}\label{eq:idm1}
   a_t = a_{max}\left[1 - \left(\frac{v_t}{\Tilde{v}}\right)^{4} - \left(\frac{\Tilde{d}}{d_t}\right)^2\right]
\end{align}
where $\Tilde{v}$ is a desired speed and $\Tilde{d}$ is the desired distance headway defined as:
\begin{align}\label{eq:idm2}
    \Tilde{d} = d_{0} + v_{t}\tau - \frac{v_t\Delta v_t}{2\sqrt{a_{max}b}}
\end{align}

The IDM has the following parameters: $\Tilde{v}$ the desired speed, $a_{max}$ the maximum acceleration rate which can be implemented by the driver, $d_0$ the minimum allowable distance headway, $\tau$ the desired headway time, and $b$ the maximum deceleration rate. To capture heterogeneity in data, we specify an ``ensemble" of IDM models, i.e. a distribution over actions given observations. The ensemble model parameters, i.e. mean and variance of action given observations, are obtained by maximizing the likelihood of the ensemble model predictions to the dataset of observed actions subject to the mean action satisfying \eqref{eq:idm1}.

\textbf{Active Inference Driving Agent:} We parameterized the state transition probability distributions $\mathbb{T}_{\hat{\theta}_1}(s'|s, a)$ as categorical distributions and the observation probability distributions $\mathbb{O}_{\hat{\theta}_1}(o|s)$ using normalizing flows \cite{papamakarios2021normalizing}, specifically, a shared inverse autoregressive flow with a set of Gaussian base distributions \cite{kingma2016improved}. Thus, observations that have high density under the conditional distribution of each state represent the ``prototypical" observation for that state. The active inference preference distribution $\Tilde{P}_{\hat{\theta}_2}(s)$ is parameterized using a categorical distribution. We then obtained the finite horizon policy in (\ref{softmax}) by computing the value function in (\ref{soft_optimal_value}) using a finite number of value iterations steps using the QMDP method \cite{littman1995learning}. By optimizing the policy log likelihood (\ref{formulation}), both the preferences and prototypical observations are fitted to explain actions in the dataset.

\vspace{-0.5cm}
\subsection{Dataset}
We trained and evaluated AIDA, BC, and IDM using the INTERACTION dataset \cite{zhan2019interaction}, a publicly available driving dataset recorded using drones on fixed road segments in the USA, Germany, and China. The dataset provides a lanelet2 format map \cite{poggenhans2018lanelet2} and a set of time-indexed trajectories of the positions, velocities, and headings of each vehicle in the scene in the map's coordinate system at a sampling frequency of 10 Hz, and the vehicle's length and width for each road segment. The dataset contains a variety of traffic behaviors, including car following, free-flow traffic, and merges. 

\begin{figure}
    \centering
\includegraphics[width=0.5\textwidth]{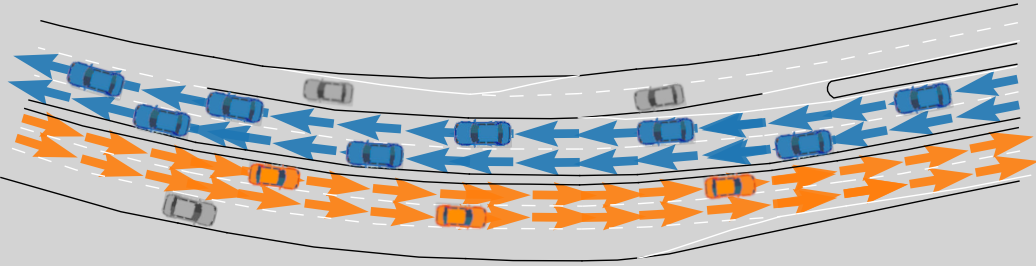}
    \caption{Top-down view of the roadway explored in the analysis. The west-bound lanes (blue) have denser traffic and more stop-and-go behavior whereas the east-bound lanes (orange) have sparser traffic and higher speed. We trained the models to emulate the behavior of the blue cars and evaluated the models’ ability to predict the behavior of the blue and orange cars. Grey cars in the merging lanes were excluded.}
    \label{fig:map}
\end{figure}

Due to our emphasis on modeling longitudinal control behavior in car following, we selected a subset of the data to include car following data from a two-way, seven-lane highway segment in China with a total distance of 175 m. We focused on vehicles in the four middle lanes shown in Figure \ref{fig:map}, where the blue west-bound lanes have denser traffic and more stop-and-go behavior and the orange east-bound lanes have sparser traffic at higher speed. We further filtered the remaining vehicles according to two criteria: 1) there was a lead vehicle with a maximum distance headway of 60 m, and 2) the ego vehicle was not performing a merge or lane change. This focus facilitates algorithm comparisons by removing environmental artifacts. We identified merging and lane change behavior using an automated logistic regression-based approach and validated the classifications with a manual review of a subset of trajectories. We also removed all trajectories with length shorter than 10 seconds for the dense lanes and 5 seconds for the sparse lanes, leaving a total of 1,254 trajectories in the dense lanes and 290 trajectories in the sparse lanes with an average length of 14 seconds. We only used the dense lane data for training models.

\subsubsection{Feature Computation}
The input features to the IDM are defined in (\ref{eq:idm1}) and (\ref{eq:idm2}). For BC and the AIDA, we used $d$ and $\Delta v$ but excluded $v$ to prevent the models from achieving spuriously high training accuracy by computing acceleration predictions from past ego velocities, a well-known phenomenon reported in prior studies \cite{de2019causal, codevilla2019exploring, zhou2017recurrent}. Furthermore, we included an additional feature $\tau^{-1}$ in BC and AIDA which is a visual estimate of inverse time-to-collision defined as the rate of change of the visual angle of the lead vehicle from the ego driver's seat position divided by the angle itself \cite{lee1976theory}. This feature was chosen to account for speed control and puts the information contained in the inputs to BC and the AIDA on a similar level to the IDM as the IDM implicitly accounts for time-to-collision in its desired distance headway computation in (\ref{eq:idm2}). It also makes our model consistent with recent family of driver models \cite{svard2021computational,engstrom2018simulating, mcdonald2019toward}.

We computed all features in the Frenet frame (i.e., lane-centric coordinates \cite{werling2010optimal}), by first transforming vehicle positions, velocities, and headings using the current lane center line as the reference path and then computing the features from the transformed positions and velocities. We obtained the drivers' instantaneous longitudinal control inputs (i.e., accelerations) from the dataset by differentiating the Frenet frame longitudinal velocities. For BC and the AIDA, we discretized the continuous control inputs into discrete actions using a Gaussian mixture model of 15 Gaussian components with mean and variance parameters chosen with the Bayesian Information Criteria \cite{murphy2012machine}.

\vspace{-0.4cm}
\subsection{Model Evaluation and Comparison}
We evaluated and compared our models' ability to generate behavior similar to the human drivers in the dataset using both open-loop offline predictions and closed-loop online simulations. In both cases, we evaluated the models (15 seeds for each model class) on two different held-out testing datasets. The first dataset includes vehicles from the same dense lanes as the training dataset. This dataset tests whether the models can generalize to unseen vehicles in the same traffic condition. We obtained this dataset by dividing trajectories in the dense lanes using a 7-3 train-test ratio. The second dataset includes vehicles from the sparse lanes. This dataset tests whether the models can generalize to unseen vehicles in novel traffic conditions, since the traffic in the east-bound lanes have on average higher speed and less density. 

\subsubsection{Offline Evaluation}\label{sec:methods_offline_eval}
The goal of the offline evaluation was to assess each model's ability to predict a driver's next action based on the observation-action history recorded in the held-out testing dataset. This task evaluates the models' ability to be used as a short-horizon predictor of other vehicles' behavior in an on-board trajectory planner \cite{sadigh2016planning}. We measured a model's predictive accuracy using Mean Absolute Error (MAE) of the predicted control inputs (unit=$m/s^2$) on the entire held-out testing datasets. For the IDM, the predicted control inputs were given by the IDM rule, i.e., we discarded the variance used for model fitting. For BC and the AIDA,  predicted action is produced by first sampling a discrete action from the action distribution predicted by the models and then sampling the mean of the selected Gaussian component from the Gaussian mixture model used to perform action discretization. MAE of each dataset action was calculated as the average of 30 samples.

\subsubsection{Online Evaluation}\label{sec:methods_online_eval}
Rather than predicting instantaneous actions, the goal of the online evaluation was to assess the models' ability to generate trajectories similar to human drivers such that they can be used as simulated agents in automated vehicle training and testing environments \cite{igl2022symphony}. This is fundamentally different from offline predictions because the models need to choose actions based on observation-action history generated by its own actions rather than those stored in the fixed, offline dataset. This can introduce significant distribution shift \cite{spencer2021feedback} sometimes resulting in situations outside the model's training data, which can lead to poor action selection. 

We built a single-agent simulator where the ego vehicle's longitudinal acceleration is controlled by the trained models and its lateral acceleration is controlled by a feedback controller for lane-centering. The lead vehicle simply plays back the trajectory recorded in the dataset. Other vehicles do not have any effect on the ego vehicle, given our observation space does not contain other vehicle related features. We tested the models on 100 randomly chosen trajectories in each of the dense-lane and sparse-lane settings.

Following \cite{suo2021trafficsim}, we measured the similarity between the generated trajectories and the true trajectories using the following metrics:
\begin{enumerate}
    \item Average deviation error (ADE; unit=$m$): deviation of the Frenet Frame position from the dataset averaged over all time steps in the trajectory.
    \item Lead vehicle collision rate (LVCR; unit=$\%$): percentage of testing trajectories containing collision events with the lead vehicle. A collision is defined as an overlap between the ego and lead vehicles' bounding boxes. 
\end{enumerate}

\subsubsection{Statistical Evaluation}
Following the recommendations in \cite{colas2019hitchhiker, agarwal2021deep} for evaluating learned control policies in stochastic environments with a finite number of testing runs, we represented the central tendency of a model's offline prediction and online control performance using the interquartile mean (IQM) of the offline MAEs and online ADEs. The IQMs are computed by 1) ranking all tested trajectories by their respective performance metrics and 2) computing the mean of the performance metrics ranked in the middle 50\%. Collision rates are computed as the percentage of testing runs that resulted in a collision. It should be noted that IQM makes the difference between each model's performance central tendency more salient at the expense of removing the tails of the performance distribution. Thus, we also provide the average performance results in appendix (\ref{sec:iqm_comparison}). To compare the central performance difference between the AIDA and baseline models, we performed two-sided Welch's t-tests with 5 percent rejection level on the MAE-IQM and ADE-IQM values computed from different random seeds with the assumption that the performance distributions between two models may have different variances \cite{colas2019hitchhiker, agarwal2021deep}.

\vspace{-0.4cm}
\subsection{Results and Discussion}
\subsubsection{Offline Performance Comparison}
\begin{figure}[!htb]
\centering
\includegraphics[width=0.4\textwidth]{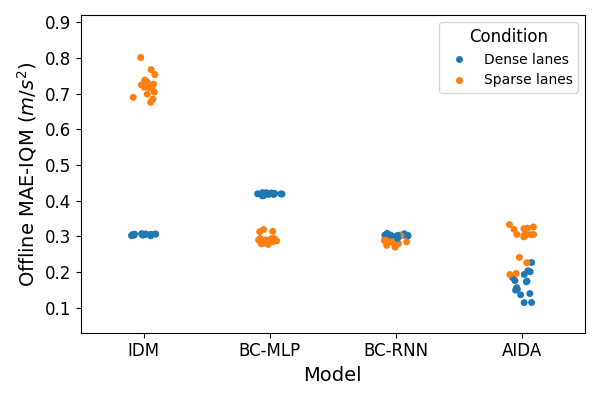}
\captionof{figure}{Offline evaluation MAE-IQM. Each point corresponds to a random seed used to initialize model training and its color corresponds to the testing condition of either dense-lane or sparse-lane.}
\label{fig:offline_mae}
\vspace{-0.7cm}
\end{figure}

Figure \ref{fig:offline_mae} shows the offline evaluation results for each model with the model type on the x-axis and the IQMs of acceleration prediction MAEs averaged across the testing dataset on the y-axis. The color of the points in the figure represents the testing condition and each point corresponds to the result of a model initialized from a different random seed. The points are randomly distributed around each x-axis label for clarity. Dispersion on the y-axis indicates sensitivity in the model to initial training conditions. The plot illustrates that the AIDA had the lowest MAE-IQM in the sparse-lane tests, followed by BC-RNN, IDM, and BC-MLP. The corresponding pairwise Welch's t-test results in Table \ref{tab:offline_eval} (Appendix \ref{sec:statistical_tests}) show that the differences between AIDA and baseline models are significant. The difference between IDM and BC-RNN was surprisingly small and BC-MLP had substantially larger MAE. This was likely because the IDM rule was well-suited to capture behavior in this traffic condition, whereas the accuracy of BC-MLP was restricted by the features it had access to and action discretization. In the sparse-lane tests, AIDA performed similarly to BC models with a few seeds substantially better than BC models. IDM performed substantially worse and also with much higher variance across different seeds. Given IDM trained from different initializations converged to similar final parameters, this result was most likely due to the distribution shift between training and testing sets and IDM rule's lack of adaptability to different traffic conditions.
However, the poor performance of IDM may be specific to the dataset considered in this paper (see Appendix, Section \ref{sec:iqm_comparison}).

\begin{figure}[!h]
        \centering        \includegraphics[width=0.4\textwidth]{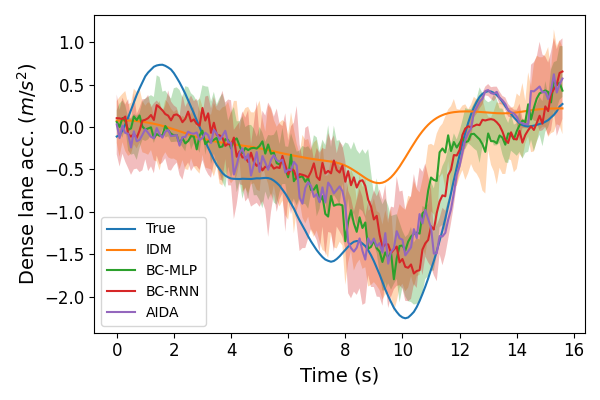}
\includegraphics[width=0.4\textwidth]{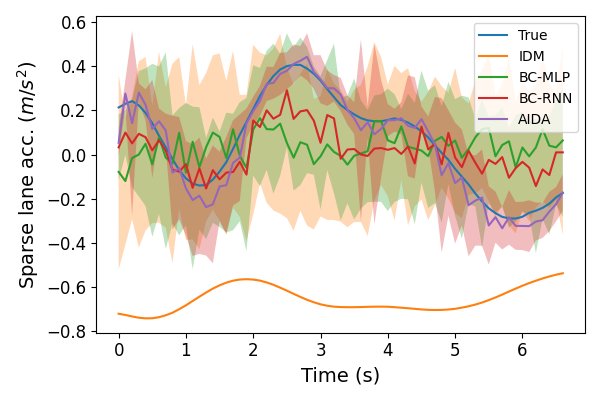}
    \caption{Example offline predictions in the dense-lane (top) and sparse-lane (bottom) settings. Each line except for the blue line represents the mean prediction of the corresponding model. Shading represents 1 standard deviation of prediction interval. The prediction intervals for BC and AIDA are computed by drawing 30 samples from the models' predictive distributions. IDM has no prediction interval because it's deterministic.}
    \label{fig:offline_pred}
\end{figure}

To understand each model's actual behavior, Figure \ref{fig:offline_pred} compares the predicted actions of each model's best performing seed versus the ground truth on a randomly selected trajectory in the dense-lane (left) and sparse-lane (right) settings, respectively, where shading corresponds to 1 standard deviation of the predictive distribution represented by 30 samples as described in section \ref{sec:methods_offline_eval}. In the dense-lane setting, all models captured the variation of actions in the dataset, i.e., acceleration first decreased and then increased. However, the acceleration magnitudes predicted by IDM were substantially smaller than the ground truth. In the sparse-lane setting, the prediction interval of all BC models and AIDA were able to cover ground truth actions. However, IDM predictions were substantially lower than the ground truth. These patterns are consistent with the aggregate measures in Figure \ref{fig:offline_mae}.

\subsubsection{Online Performance Comparison}
\begin{figure}[!htb]
\centering
\includegraphics[width=0.4\textwidth]{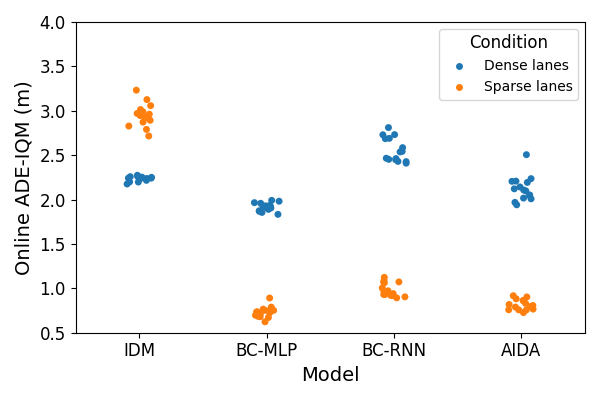}
\captionof{figure}{Online evaluation ADE-IQM. Each point corresponds to a random seed used to initialize model training and its color corresponds to the testing condition of either dense-lane or sparse-lane.}
\label{fig:online_mae}
\end{figure}

Figure \ref{fig:online_mae} shows the IQM of each model's ADEs from data set trajectories in the online evaluations using the same format as the offline evaluation results. In the dense-lane testing condition, all models had ADE-IQM values between 1.8 m and 2.8 m, which is less than the length of a standard sedan ($\approx$ 4.8 m; \cite{sedan}). Among all models, BC-MLP achieved the lowest ADE values for both the dense-lane and sparse-lane conditions, followed by the AIDA, IDM, and BC-RNN. Furthermore, both the AIDA and BC models achieved lower ADE-IQM in the sparse lane settings compared to the dense-lane setting, however the IDM achieved higher ADE-IQM in the sparse-lane setting. The Welch's t-test results in Table \ref{tab:online_eval} show that AIDA's online test performances are significantly different from all baseline models in both the dense-lane and sparse-lane settings (P $\leq$ 0.01). These findings confirm that the AIDA and BC models generalized better to the sparse-lane setting than the IDM and suggest that the AIDA's average online trajectory-matching ability is {\em on average} better than IDM and BC-RNN, although BC-MLP is better than the AIDA. However, it should be noted that the tail-end behavior of AIDA and BC-RNN can be worse when evaluated under average ADE (i.e., without IQM; see Figure \ref{fig:online_ade_iqm_comparison} in Appendix \ref{sec:iqm_comparison}) where the worst AIDA seed performed approximately equal to the worst BC-RNN seed, both of which would increase online ADE by 1 m.

\begin{figure}[!h]
\centering
\includegraphics[width=0.4\textwidth]{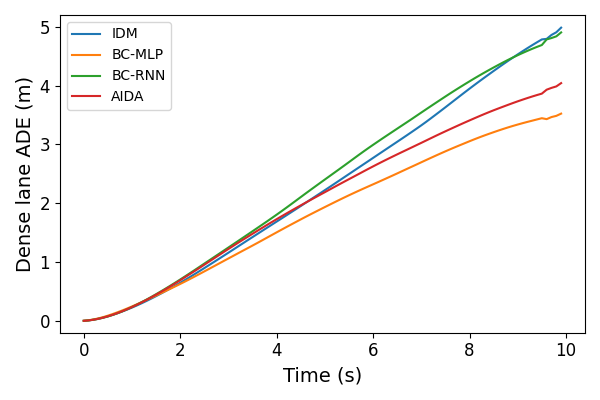}
\includegraphics[width=0.4\textwidth]{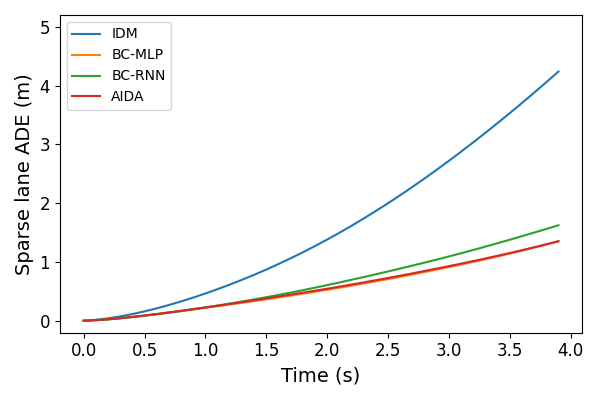}
    \caption{Online evaluation ADE for each time step averaged over all online testing episodes for the dense-lane (top) and sparse-lane (bottom) settings by the best seed of each model.}
    \label{fig:online_pred}
\end{figure}

To understand how trajectory deviations were generated, Figure \ref{fig:online_pred} shows the ADE of the best seed of each model averaged over all testing episodes for each time step in the dense-lane (left) and sparse-lane (right) scenarios. We truncated the plots at 10 s and 4 s because there are very few trajectories longer than those horizons making the curves highly oscillatory. The amount of deviations generated by different models are consistent with the prior study \cite{bhattacharyya2020modeling}. The ranking of model performance is also consistent with the aggregated measures in Figure \ref{fig:online_mae}. In the dense-lane settings (Figure \ref{fig:online_pred} left), model performance started to differentiate around 4 s but the differences were not substantial (i.e., up to 2 m). In contrast, in the sparse-lane setting, IDM generated substantially larger deviations from the beginning and BC-MLP and AIDA had nearly matching ADE at all time steps.

\begin{figure}[!htb]
\centering
\includegraphics[width=0.4\textwidth]{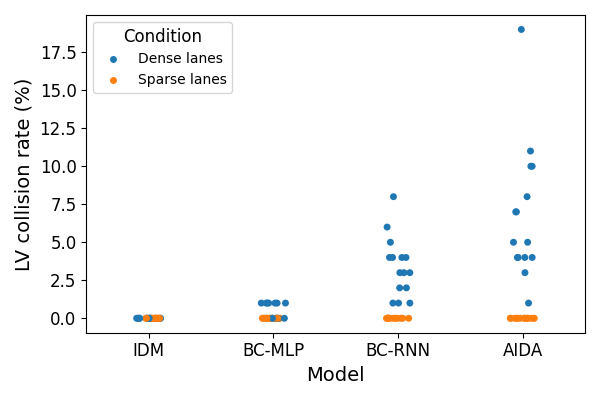}
    \captionof{figure}{Lead vehicle collision rate in online evaluation. Each point corresponds to a random seed used to initialize model training and its color corresponds to the testing condition of either dense-lane or sparse-lane.}
    \label{fig:online_cr_lv}
\end{figure}

Figure \ref{fig:online_cr_lv} shows the lead vehicle collision rates for each random seed and model using the same format as Figure \ref{fig:online_mae}. The figure illustrates that in the dense-lane condition, the random seeds for BC-MLP, BC-RNN, and the AIDA had more collisions than the IDM (0\% collision rate across all seeds). In particular, BC-RNN and the AIDA had substantial differences across random seeds compared to the other models. However, the minimum collision rates for BC-MLP, BC-RNN, and the AIDA were consistent (less than or equal to 1\%). In the sparse-lane condition, the collision rate was 0\% for all four models. The higher collision rates in the dense-lane data are likely due to the traffic density and complexity, which were higher in the dense-lane condition compared to the sparse-lane condition. This is also due to the way we defined a collision in section \ref{sec:methods_online_eval} as any overlapping of vehicle bounding boxes. As we show later in section \ref{sec:model_diagonostics}, many collision events were due to insufficient braking magnitude despite correct braking intent, part of which can be attributed to discrete belief and action spaces. This puts ego vehicle's stopping position slightly ahead of the no-collision position without generating a large position deviation as commonly seen in machine-learned driving agents \cite{bhattacharyya2020modeling}.

\subsubsection{AIDA Interpretability Analysis}
The previous sections suggest that the AIDA can capture driver car following behaviorcomparably if not better than baseline models. However, the findings have yet addressed the interpretability of the AIDA. Interpretability represents the ability to understand the relationship between model input and output and is a crucial element of model deployment success \cite{rudin2019stop}. While there is no established metric for model interpretability, R{\"a}ukur et. al. \cite{raukur2022toward} recommend assessments based on the ease of comprehending the connection between model input and output and tracing model predictive errors to internal model dynamics. Given that the AIDA's decisions are emitted from a two-step process, i.e., (1) forming beliefs about the environment and (2) selecting control actions that realized preferred states (i.e., minimize free energy), the model's interpretability depends on the two sub-processes both independently and jointly. Thus, we examined the learned input-output mechanism by visualizing the components (i.e., the observation, transition, and preference distributions) of the best performing AIDA seed and verified them against expectations guided by driving theory \cite{fuller2005towards, engstrom2018great,summala1997hierarchical}. We then examined the joint belief-action process by replaying the AIDA beliefs and diagnosing its predictions of recorded human drivers in the offline setting and its own decisions in the online setting. 

\begin{figure}[!htb]
\centering
    \begin{subfigure}[b]{0.22\linewidth}
    \includegraphics[height=0.35\textheight, width=1\linewidth, trim={0 4.5cm 0 0}, clip]{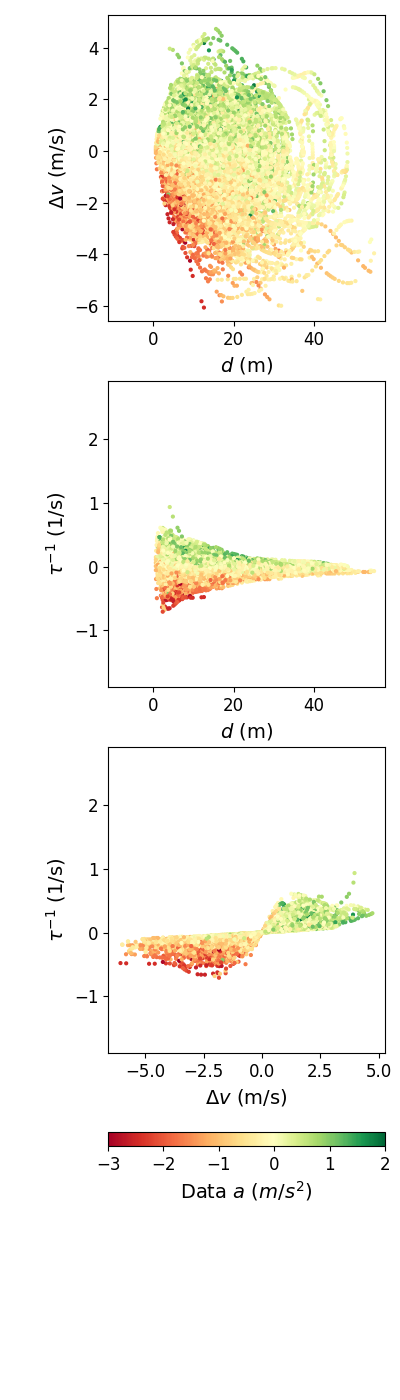}
    \caption{}
    \label{fig:obs_scatter_data}
    \end{subfigure}
    \begin{subfigure}[b]{0.22\linewidth}
    \includegraphics[height=0.35\textheight, width=1\linewidth, trim={0 4.5cm 0 0}, clip]{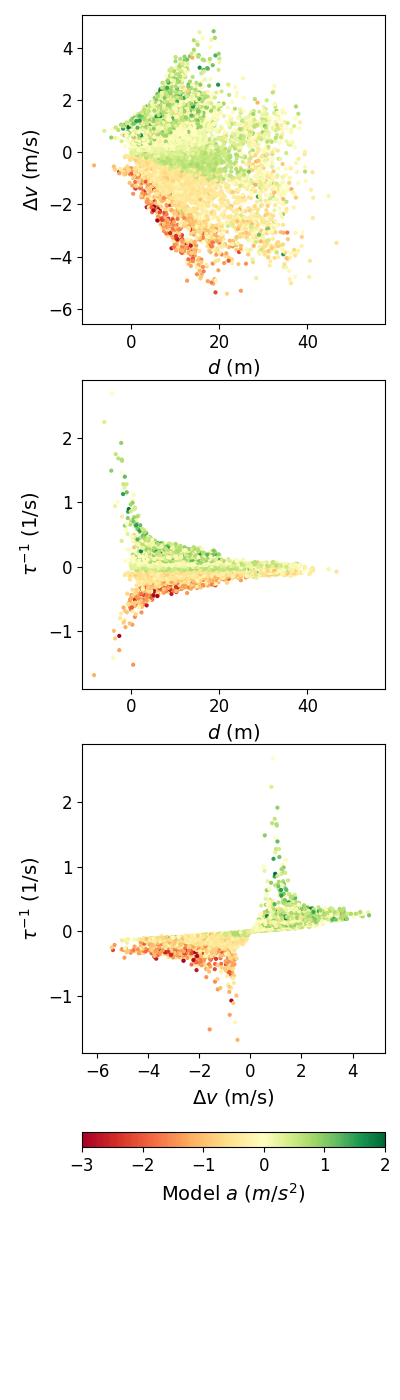}
    \caption{}
    \label{fig:obs_scatter_policy}
    \end{subfigure}
    \begin{subfigure}[b]{0.22\linewidth}
    \includegraphics[height=0.35\textheight, width=1\linewidth, trim={0 4.5cm 0 0}, clip]{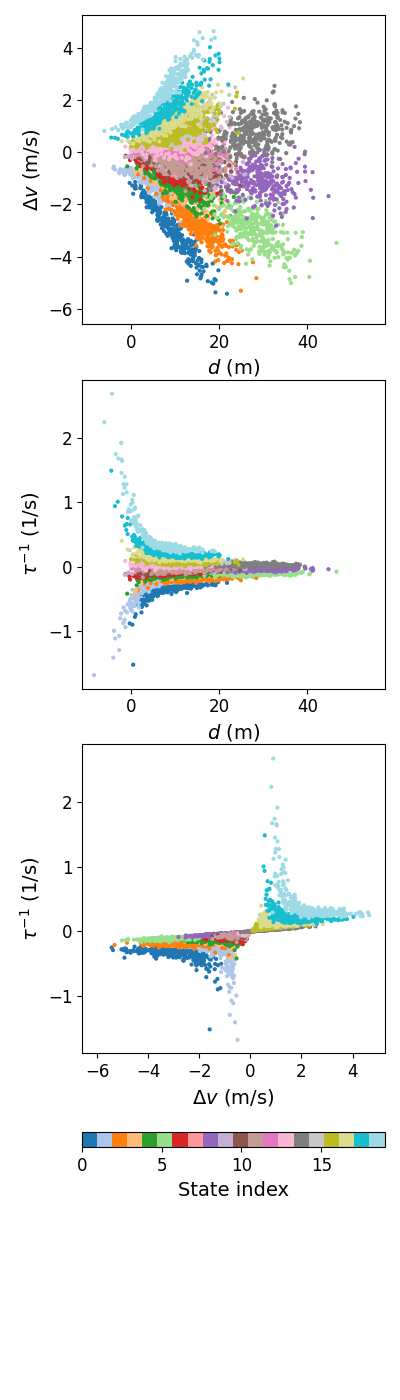}
    \caption{}
    \label{fig:obs_scatter_cluster}
    \end{subfigure}
    \begin{subfigure}[b]{0.22\linewidth}
    \includegraphics[height=0.35\textheight, width=1\linewidth, trim={0 4.5cm 0 0}, clip]{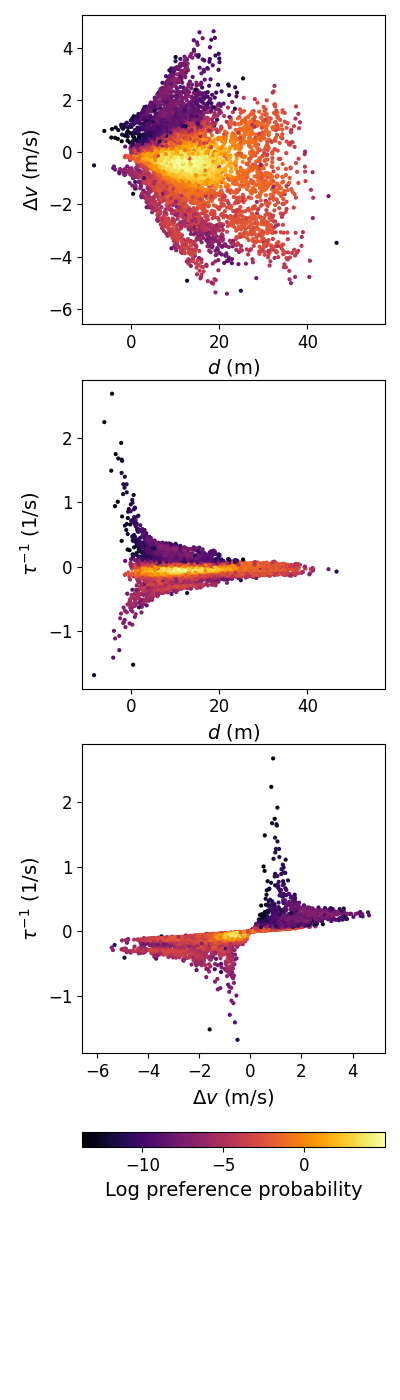}
    \caption{}
    \label{fig:obs_scatter_preference}
    \end{subfigure}
\caption{Visualizations of the best performing AIDA seed. In panel (a), we plotted observations sampled from the dataset. In panels (b), (c), and (d)  we illustrate AIDA's learned policy, observation model, and preference model via 200 ``prototypical" samples from each state's conditional observation distribution $\mathbb{O}(o|s)$ and plotted the samples for each pair of observation feature combinations. The points in each panel are colored by: (a)  accelerations from the dataset, (b) predicted accelerations upon observing the sampled signals from a uniform prior belief, (c) state indices (d) log probabilities of the preference distribution.}
\label{fig:obs_scatter}
\vspace{-0.6cm}
\end{figure}

\subsubsection{AIDA Component Interpretability}
Initial insights into the model input and output connections can be gained by visualizing the AIDA components, specifically its policy (Figure \ref{fig:obs_scatter_policy}), observation distribution (Figure \ref{fig:obs_scatter_cluster}), and preference distribution (Figure \ref{fig:obs_scatter_preference}). These figures show 200 random ``prototypical" samples from the observation distribution $\mathbb{O}(o|s)$ of each state, plotted on each pair of observation modalities. The top row shows the samples using distance headway ($d$; x-axis) by relative velocity to the lead vehicle ($\Delta v$; y-axis), the middle row shows distance headway by $\tau^{-1}$, and the bottom row shows relative velocity by $\tau^{-1}$. Color is used to highlight relevant quantities of interest. We further used samples drawn from the INTERACTION dataset, plotted in Figure \ref{fig:obs_scatter_data} and colored by the recorded accelerations, to facilitate interpreting the the AIDA samples. The shape of the sampled points matches the contour of the empirical dataset (Figure \ref{fig:obs_scatter_data}), particularly in the middle and bottom visualizations, which suggests that the model's learned observation model aligns with the recorded observations in the dataset. However, the learned distributions also showed longer tails at the edge of the data distribution. This was expected because the dataset does not contain samples that correspond to these extreme conditions. Thus the model could not learn accurate kinematics in these regions. Nevertheless, this does not affect the interpretability analysis.

Figure \ref{fig:obs_scatter_policy} illustrates the observation samples by the model's chosen control actions. Darker green and red colors correspond to larger acceleration and deceleration magnitudes, respectively, and light yellow color corresponds to near zero control inputs. The color gradient at different regions in Figure \ref{fig:obs_scatter_policy} is consistent with that of the empirical dataset shown in Figure \ref{fig:obs_scatter_data}. This shows that the model learned a similar control rule (i.e., observation to action mapping) as the empirical dataset. The control rule can be interpreted as the tendency to choose negative accelerations when the relative speed and $\tau^{-1}$ are negative and the distance headway is small, and positive accelerations in the opposite case. Furthermore, the sensitivity of the red and green color gradients with respect to distance headway shows that the model tends to accelerate whenever there is positive relative velocity, regardless of the distance headway. However, it tends to input smaller deceleration at large distance headway for the same level of relative speed. 

Figure \ref{fig:obs_scatter_cluster} shows the observation samples colored by their associated discrete states. The juxtaposition of color clusters in the top panel shows that the AIDA learned to categorize observations by relative speed and distance headway and its categorization for relative speed is more fine-grained at small distance headways and spans a larger range of values. The middle and bottom panels show that its categorization of relative speed is highly correlated with $\tau^{-1}$ as the ordering of colors along the y-axis is approximately the same as in the top panel. The middle and bottom panels show that the AIDA's categorization of high $\tau_1$ magnitude states (blue and cyan clusters) have a larger span than that of low $\tau^{-1}$ magnitude states. These patterns further establish that the AIDA has learned a representation of the environment consistent with the dataset. At the same time, it can be interpreted as a form of satisficing in that the model represents low urgency large distance headway states with less granularity \cite{hancock1999car}. 

Figure \ref{fig:obs_scatter_preference} shows the observation samples by the log of its preference probability, $\Tilde{P}(o) = \sum_{s}\Tilde{P}(s)\mathbb{O}(o|s)$, where higher preference probability (i.e., desirability) corresponds to brighter colors (e.g., yellow) and lower desirability corresponds to darker colors (e.g., purple). The figure shows that the highest preference probability corresponds to observations of zero $\tau^{-1}$, zero relative velocity, and a distance headway of 18 m (see the center region of the middle chart, and the yellow circle at the left-center of the top chart). This aligns with the task-difficulty homeostasis hypothesis that drivers prefer states in which the crash risk is manageable \cite{fuller2005towards} and not increasing. It is also consistent with the observed driver behavior in Figure \ref{fig:obs_scatter_data} where drivers tend to maintain low accelerations (light yellow points) within the same regions. 

Overall, these results show a clear mapping between the AIDA's perceptual (Figure \ref{fig:obs_scatter_cluster}) and control (Figure \ref{fig:obs_scatter_preference} and \ref{fig:obs_scatter_policy}) behavior that is both consistent with the observed data and straightforwardly illustrated using samples from the fitted model distributions. This mapping facilitates predictions of the AIDA's reaction to observations without querying the model, which is an important dimension of interpretability in real world model validation \cite{raukur2022toward}.

\subsubsection{AIDA Decision Diagnostics}\label{sec:model_diagonostics}
While the previous analysis illustrates the interpretability of individual model components, the overall model interpretability is also contingent upon understanding the interaction between components. To address this, we analyzed two dense-lane scenarios where the AIDA made sub-optimal decisions in the model testing phase --- one from the offline evaluations where the AIDA's predictions had the largest MAE and one from the online evaluations where the AIDA generated a rear-end collision with the lead vehicle. We first visualized the AIDA's beliefs and policies as the model generated actions and then used those visualizations to demonstrate how the transparent input-output mechanism in the AIDA can be used to mitigate the sub-optimal decisions. 

\begin{figure}[!htb]
\centering
    \begin{subfigure}[b]{0.8\linewidth}
    \includegraphics[width=1\linewidth]{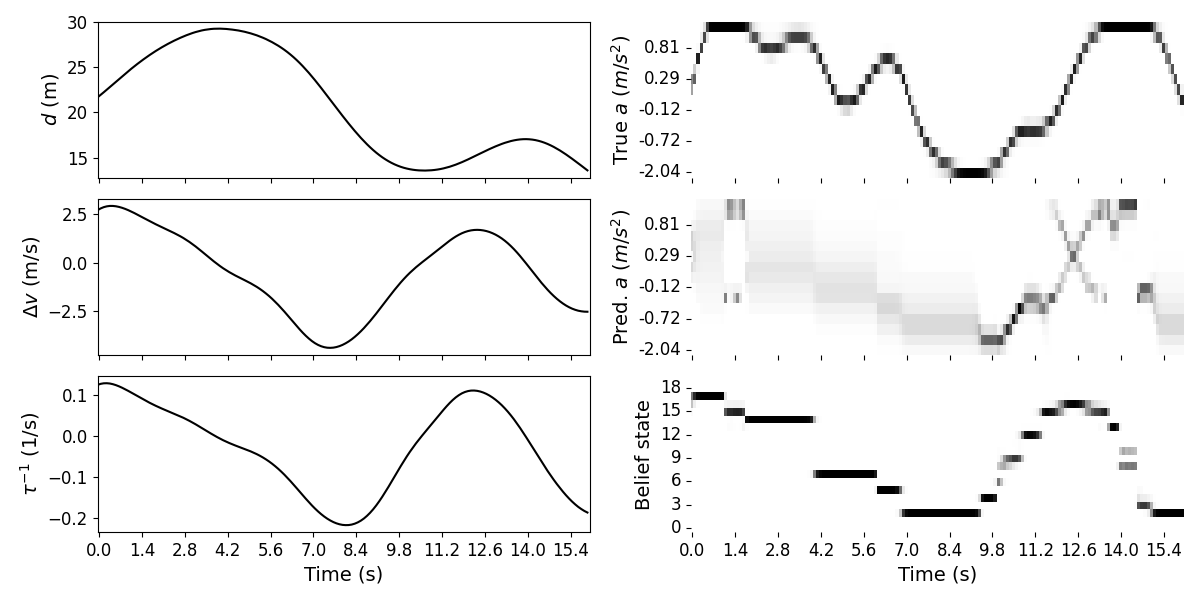}
    \end{subfigure}
\caption{Visualizations of a dense-lane offline evaluation trajectory where the AIDA had the highest prediction MAE. The charts in the left column show distance headway, relative speed, and $\tau^{-1}$ signals observed by the model over time. The binary heat maps in the right column show the ground truth action probabilities (top), action probabilities predicted by the AIDA (middle), and the corresponding belief states (bottom) over time (x-axis), where darker colors correspond to higher probabilities. The belief state and action indices are sorted by the mean $\tau^{-1}$ and acceleration value of each state, respectively.}
\label{fig:offline_traj}
\end{figure}

The chosen offline evaluation trajectory is visualized in Figure \ref{fig:offline_traj}. The left column charts show the data of the three observation features over time. The right column charts show the time-varying ground truth action probabilities over time (top), action probabilities predicted by the AIDA over time (middle), and environment state probabilities $P(s|h)$ inferred by the AIDA over time (bottom). In the right-middle and right-bottom charts, the action and belief state indices are sorted by the mean acceleration and $\tau^{-1}$ value of each state to facilitate alignment with the left and top-right charts. We labeled the actions by the corresponding means but not the belief states because they represent multi-dimensional observation categorizations (see Figure \ref{fig:obs_scatter_cluster}). The bottom-right chart shows that the inferred belief patterns closely followed the observed relative speed and $\tau^{-1}$ in the left-middle and left-bottom charts with high precision, i.e., close to probability of 1. The predicted action probabilities in the right-middle chart followed the trend of the ground truth actions, however, they exhibited substantially higher uncertainty at most time steps and multi-modality at $t=1 \text{ s}$ and $t=12 \text{ s}$, where one of the predicted modes coincided with the true actions. Given the inferred beliefs were precise, uncertain and multi-model actions were likely caused by inter-driver variability in the dataset, where drivers experienced similar belief states but selected different actions. Alternatively, this uncertainty may be caused by actual drivers having highly different beliefs after experiencing similar observations, 
In either case, the error in AIDA predictions can be attributed to inconsistency between the belief trajectories and action predictions.

\begin{figure}[!htb]
\centering
    \begin{subfigure}[b]{0.8\linewidth}
    \begin{tikzpicture}
    \node[anchor=south west,inner sep=0] (image) at (0,0) {\includegraphics[width=1\textwidth]{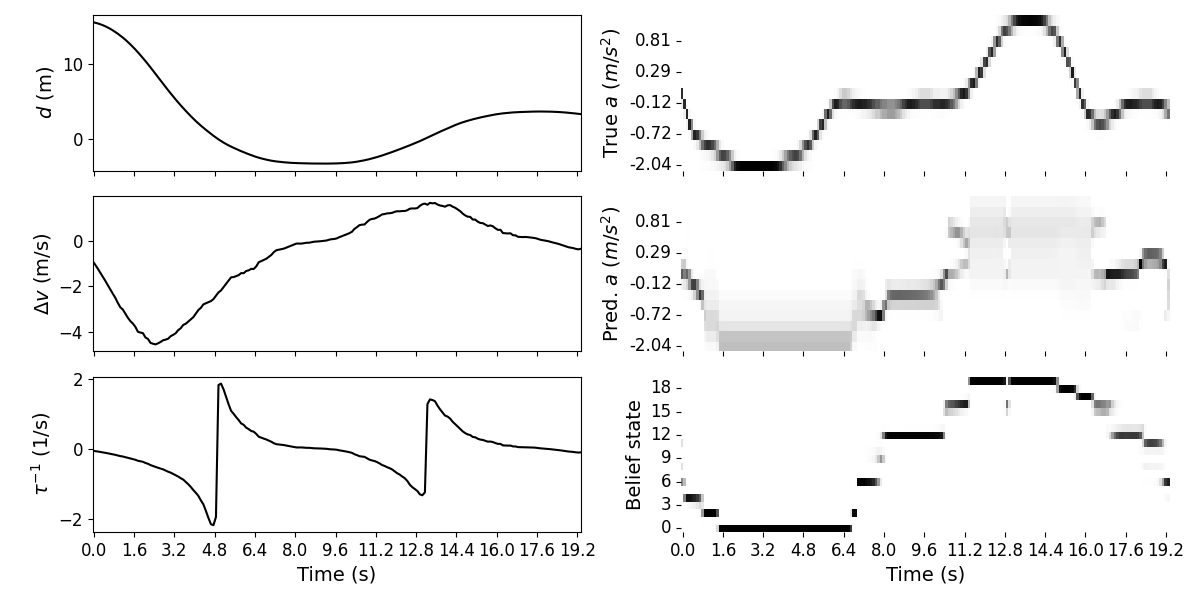}};
    \begin{scope}[x={(image.south east)},y={(image.north west)}]
        \draw[red,ultra thick,rounded corners] (0.17,0.12) rectangle (0.37,0.36);
    \end{scope}
    \end{tikzpicture}
    \end{subfigure}
\caption{Visualizations of a dense-lane online evaluation trajectory where the AIDA generated a rear-end collision with the lead vehicle. This figure shares the same format as Figure \ref{fig:offline_traj}. The red square in the bottom-left chart represents the duration of the rear-end crash event where the vehicle controlled by the AIDA had overlapping bounding box with the lead vehicle.}
\label{fig:online_traj}
\end{figure}

The chosen online evaluation trajectory which resulted in a rear-end collision with the lead vehicle is shown in Figure \ref{fig:online_traj} plotted using the same format as Figure \ref{fig:offline_traj}. The duration of the crash event is highlighted by the red square in the bottom-left chart, where the sign of $\tau^{-1}$ values instantly inverted when overlapping bounding boxes between the ego and lead vehicle first occurred and eventually ended. The AIDA initially made the correct and precise decision of braking, however, its predictions for high magnitude actions became substantially less precise prior to the collision ($t > 1 \text{ s}$; see right middle chart). This led to the model failing to stop fully before colliding with the lead vehicle. The belief pattern shows that the AIDA tracked the initial decreasing values of relative speed and $\tau^{-1}$ but did not further respond to increasing magnitude of $\tau^{-1}$ 3 seconds prior to the crash (starting at $t = 1.6 \text{ s}$). These findings show that the model exhibited the correct behavior of being ``shocked" by out-of-sample near-crash observations, however, the learned categorical belief representation was not able to extrapolate beyond the data from the crash-free INTERACTION dataset.

The analysis of the near-crash AIDA beliefs suggests that editing the AIDA's learned environment dynamics model (i.e., the transition and observation distributions) to properly recognize near-crash observation signals can likely avoid the current crash.

The analyses in this section show that the decision making structure in the AIDA enables modelers to reason about the training dataset's effect on the learned model behavior. To the best of our knowledge, this analysis is not possible with neural network BC models using existing interpretability tools. Thus AIDA represents a significant step forward for interpretable perception and control models of human control behavior.

\section{Conclusions}
We consider the problem of learning a model of human perception and control based on data in the form of observations and implemented actions.
We posit a POMDP model and formulated a bi-level optimization formulation of Maximum A Posteriori (MAP) estimate for the primitives of the model.
To illustrate the estimation methodology we develop
a model of driver behavior (AIDA) with the reward specification motivated by the active inference framework from cognitive science.
Using car following data, we showed that the AIDA performed comparably and in certain cases better than the rule-based IDM and data-driven neural network benchmarks. Using an interpretability analysis, we showed that the structure of the AIDA provides superior transparency of its input-output mechanics than the neural network models. Future work should focus on training with data from more diverse driving environments and examining model extensions that can capture heterogeneity across human agents.

\section*{Acknowledgements}
Support for this research was provided in part by the U.S. Department of Transportation (DOT), University Transportation Centers Program to the Safety through Disruption University Transportation Center (451453-19C36) and the UK Engineering and Physical Sciences Research Council (EPSRC; EP/S005056/1). Thanks to advisers, J. Engstrom and M. O’Kelly, from Waymo, who helped set the technical direction, identified relevant published research, and advised on the scope and structuring of this publication, independent of the support this research received from USDOT.

\section*{References}
\bibliographystyle{ieeetr}
\bibliography{ref.bib}

\section{Appendix}
\subsection{Proofs}
\subsubsection{Proof of Proposition 1}

\begin{proof}
The proof is by induction. Assume $U_{t+1}(h_{t+1}) = V_{t+1}(b_{t+1})$, then
\begin{align*}
 U_{t}(h_t)  &= \max_{\pi(\cdot|h_t)} \Bigg\{ \sum_{a} \sum_s r(s, a)\mathbb{P}(s_t=s|h_t)\pi(a|h_t) - c((\pi(\cdot|h_t))
 \\ 
 &+ \gamma \sum_{a}\sum_{o_{t+1}} \mathbb{P}(o_{t+1}|h_t, a)\pi(a|h_t) V_{t+1}(b_{t+1}) \Bigg \} \\
  &=  \max_{\pi(\cdot|b_t)} \Bigg\{ \sum_{a} r(s, a)b_t(s)\pi(a|b_t) - c((\pi(\cdot|b_t)) \\ 
  & + \gamma\sum_{a}\sum_{o_{t+1}} \sigma(o_{t+1}|s_t, a)\pi(a|b_t) V_{t+1}(b_{t+1}) \Bigg \}\\
  &=V_{t}(b_t)
\end{align*}
where $b_{t+1}(s)= P(s_{t+1}=s|h_t \cup \{a,o_{t+1}\})$ and the second equality follows from 
\begin{align*} 
\mathbb{P}(o_{t+1}|h_t, a) & = \sum_{s_t}\sum_{s_{t+1}}\mathbb{O}(o_{t+1}|s_{t+1})\mathbb{T}( s_{t+1}|s_t, a)b_t(s_t) \\
&= \sigma(o_{t+1}|b_t, a)
\end{align*}
\end{proof}

\vspace{-1cm}
\subsection{Proof of Theorem 1}

To prove \emph{(a)}, let $Q_{1},Q_{2}\in \mathcal{Q}$ and $\epsilon =\Vert
Q_{1}-Q_{2}\Vert $. Then 
\begin{align*}
\log \left( \sum_{a}\pi ^{0}(a|b)\exp \big(\frac{1}{\alpha}Q_{1}(b,a)\big)\right) & \nonumber \\ \leq
\log \left( \sum_{a}\pi ^{0}(a|b)\exp \big(\frac{1}{\alpha}Q_{2}(b,a)+\epsilon \big)\right) 
&  \\
 =\log \left( \exp (\epsilon )\sum_{a}\pi ^{0}(a|b)\exp \big(\frac{1}{\alpha}Q_{2}(b,a)\big)%
\right)    &  \\
 =\epsilon +\log \left( \sum_{a}\pi ^{0}(a|b)\exp \big(\frac{1}{\alpha}Q_{2}(b,a)\big)%
\right) 
\end{align*}%
Similarly, we have 
\begin{align*}\log \left( \sum_{a}\pi ^{0}(a|b)\exp \big(\frac{1}{\alpha}Q_{1}(b,a)%
\big)\right) & \\
\geq -\epsilon +\log \left( \sum_{a}\pi ^{0}(a|b)\exp \big(%
\frac{1}{\alpha}Q_{2}(b,a)\big)\right) &
\end{align*}
Hence, we obtain that 
\begin{equation*}
\Vert \mathcal{B}Q_{1}-\mathcal{B}Q_{2}\Vert \leq \gamma \Vert
Q_{1}-Q_{2}\Vert =\gamma \epsilon .
\end{equation*}

To prove \emph{(b)},
consider the policy of the form: 
\begin{equation*}
\pi^*(a|b)=\frac{\pi ^{0}(a|b)\exp\big(\frac{1}{\alpha}Q^*(b,a)\big)}{\sum_{a^{\prime }\in A}\pi
^{0}(a^{\prime }|b)\exp \big(\frac{1}{\alpha}Q^*(b,a^{\prime })\big)}.
\end{equation*}%
where $Q^*$ is the unique fixed point of $\mathcal{B}$.
We first note that 
\begin{align*}
\log \pi^* (a|b)&=\log \pi ^{0}(a|b)+\frac{1}{\alpha}Q^*(b,a)\\
&-\log
\sum_{a^{\prime }}\pi ^{0}(a^{\prime }|b)\exp \big(\frac{1}{\alpha}Q^*(b,a^{\prime })\big) 
\end{align*}
Thus,
\begin{align}
\frac{1}{\alpha}\Big[\sum_{a}Q^*(b,a)\pi^* (a|b)-\alpha \mathcal{D}_{KL}(\pi^* (\cdot |b)||\pi ^{0}(\cdot
|b))\Big]& \nonumber \\
=\sum_{a}\pi^* (a|b)\Big[\log \sum_{a^{\prime }}\pi ^{0}(a^{\prime }|b)\exp\big(\frac{1}{\alpha}
Q^*(b,a^{\prime })\big)+\log \frac{\pi^* (a|b)}{\pi ^{0}(a|b)}\Big] & \nonumber \\
-\mathcal{D}_{KL}(\pi^* (\cdot |b)||\pi ^{0}(\cdot |b)) & \nonumber \\
=\log \sum_{a^{\prime }}\pi ^{0}(a^{\prime }|b)\exp \big(\frac{1}{\alpha}Q^*(b,a^{\prime })\big).
\label{max_2}
\end{align}%
Moreover, for any policy $\pi\neq \pi^* $, it holds that 
\begin{align}
\frac{1}{\alpha}\Big[\sum_{a}Q^*(b,a)\pi(a|b)-\alpha\mathcal{D}_{KL}(\pi(\cdot |b)||\pi
^{0}(\cdot |b))\Big]& \nonumber \\
=\sum_{a}\pi(a|b)[\frac{1}{\alpha}Q^*(b,a)-\log \frac{\pi(a|b)}{%
\pi ^{0}(a|b)}]  & \nonumber \\
 =\sum_{a}\pi(a|b)[\log \frac{\pi^* (a|b)}{\pi ^{0}(a|b)} & \nonumber \\
 +\log
\sum_{a^{\prime }}\pi ^{0}(a^{\prime }|b)\exp \big(\frac{1}{\alpha}Q^*(b,a^{\prime })\big)-\log \frac{\pi(a|b)}{\pi ^{0}(a|b)}]  & \nonumber \\ =-D_{KL}(\pi(\cdot |b)||\pi^* (\cdot |b))+\log \sum_{a^{\prime }}\pi
^{0}(a^{\prime }|b)\exp\big(\frac{1}{\alpha} Q^*(b,a^{\prime })\big)  \label{max_1}
\end{align}%
Since $D_{KL}(\pi (\cdot |b)||\pi^*(\cdot |b))\geq 0$ we conclude from %
\eqref{max_2} and \eqref{max_1} that 
\begin{align*}
\alpha \log \sum_{a^{\prime }}\pi ^{0}(a^{\prime }|b)\exp \big(\frac{1}{\alpha}Q^*(b,a^{\prime })\big)&  \\
=\sum_{a}Q^*(b,a)\pi^* (a|b)-\alpha\mathcal{D}_{KL}(\pi^* (\cdot |b)||\pi ^{0}(\cdot |b))
&  \\
 =\max_{\pi(\cdot |b)}[\sum_{a}Q^*(b,a)\pi(a|b)-\alpha \mathcal{D}_{KL}(%
\pi(\cdot |b)||\pi ^{0}(\cdot |b))] = V^*(b)
\end{align*}%

To prove \emph{(c)}, we apply \eqref{max_2} and \eqref{max_1} to $\pi $ to
conclude that 
\begin{align*}
V^*(b) =\max_{\pi (\cdot |b)}[\sum_{a}Q^*(b,a)\pi (a|b)-\alpha \mathcal{D}_{KL}(\pi
(\cdot |b)||\pi ^{0}(\cdot |b))] & \\
=\max_{\pi (\cdot |b)}[\sum_{a}\sum_{s}r(s,a)b(s)\pi (a|b)-\alpha \mathcal{D}%
_{KL}(\pi (\cdot |b)||\pi ^{0}(\cdot |b))) & \\
+\gamma \sum_{a}\sum_{o^{\prime
}}\sigma (o^{\prime }|b,a)\pi (a|b)V^*(b^{\prime })]
\end{align*}%
where $b'$ is the updated Bayes belief distribution after action $a$ is implemented and observation $o'$ is recorded and the
optimality of $\pi^*$ follows from Proposition 1.

\subsection{Implementation details}
\label{appx:implementation}
The source code is available at \url{https://github.com/ran-weii/interactive_inference}.
\subsubsection{BC Implementation}\label{sec:appendix_bc}
For BC-MLP, we used a two-layer MLP network with ReLU activation and 40 hidden units in each layer. For BC-RNN, we used a two-layer MLP network on top of a single-layer GRU network with ReLU activation and 30 hidden units in each layer. The GRU layer only takes in past observations but not past actions. We found that larger number of hidden units in the BC-RNN model led to significant overfitting. Both BC-MLP and BC-RNN receive 3 input observations and output probability distributions over 15 discrete actions. 

\subsubsection{AIDA Implementation}\label{sec:appendix_aida}
The AIDA implementation follows the value-iteration network and QMDP network \cite{tamar2016value, karkus2017qmdp} to enable end-to-end training in Pytorch \cite{paszke2019pytorch}. We used a state dimension of 20, action dimension of 15, and a planning horizon of 30 steps (3 seconds). Discrete state transition probabilities are parameterized using categorical distributions. The continuous observation distributions are parameterized using a set of Gaussian distributions, one for each discrete state, and a shared noramlizing flow network to transform the base Gaussian distributions into more flexible density estimators. Specifically, we use inverse autoregressive flow \cite{kingma2016improved} parameterized by a two-layer MLP network with ReLU activation and 30 hidden units in each layer. 

For each mini-batch of observation-action sequences, we first computed the likelihood of the observations at all time steps and compute the belief at each time step as:
\begin{align}
    b(s_{t}) = \frac{P(o_{t}|s_{t})\sum_{s_{t-1}}P(s_{t}|s_{t-1}, a_{t-1})b(s_{t-1})}{\sum_{s_{t}}P(o_{t}|s_{t})\sum_{s_{t-1}}P(s_{t}|s_{t-1}, a_{t-1})b(s_{t-1})}.
\end{align}
We then computed the value function (\ref{soft_optimal_value}) for the EFE reward and the resulting optimal policy in (\ref{receding_soft_policy}) for each inferred belief using the QMDP approximation method \cite{littman1995learning}. The QMDP method assumes the belief-action value can be approximated as a weighted-average of the state-action value:
\begin{align}
    Q^{*}(b_t, a_t) = \sum_{s_t}b_t(s_t)\mathcal{Q}^{*}(s_t, a_t),
\end{align}
where
\begin{align}
    Q^{*}(s_t, a_t) &= r(s_t, a_t) + \log \pi(a_t|s_t) + \nonumber \\
    &+\sum_{s_{t+1}}P(s_{t+1}|s_t, a_t)V^{*}(s_{t+1}) ,\\
    r(s_t, a_t) &= EFE(s_t, a_t) = D_{KL}(P(s_{t+1}|s_t, a_t) || \Tilde{P}(s_{t+1})) \nonumber \\
    &+ \mathbb{E}_{P(s_{t+1}|s_t, a_t)}[\mathcal{H}(P(o_{t+1}|s_{t+1}))] .\label{eq:approx_efe}
\end{align}
and $\forall s \in \mathcal{S}, Q^{*}(s_{t+H+1}) = 0$. We further approximate the observation entropy using the entropy of the Gaussian base distributions of the normalizing flows which can be computed in closed form.

The combination of QMDP approximation and computing the observation entropy in (\ref{eq:approx_efe}) using the Gaussian base distributions reduced the model's ability to evaluate state uncertainty. However, given the low state uncertainty shown in Figure \ref{fig:offline_traj} and Figure \ref{fig:online_traj} (i.e., the nearly deterministic belief states in the lower right charts), these approximations do not significantly impact the current results while providing the benefit of computational tractability. 

\subsubsection{Training and hyperparameters}
We trained all models using the Adam optimizer for a fixed number of epochs which are selected upon visual inspection of convergence, i.e., the loss function no longer changes significantly. For all models, we use a batch size of 100. For AIDA, we use a smaller learning rate of 0.001 for the normalizing flow network than the rest of the model because of its sensitivity to large learning rates. Additional hyperparameters are reported in table \ref{hyperparams}. 

\begin{table}[!htb]
\caption{Training hyperparameters}
\label{hyperparams}
\centering
\begin{tabular}{ccccc}
\hline
Hyperparameter & IDM & BC-MLP & BC-RNN & AIDA \\
\hline
Learning rate & 0.005 & 0.001 & 0.001 & 0.01 \\
Training epochs & 300 & 500 & 500 & 500\\
\hline
\end{tabular}
\end{table}

\begin{table}[!htb]
\caption{Model input features}\label{tab:model_features}
\centering
\begin{tabular}{lcccc}
\hline
Feature & IDM & BC-MLP & BC-RNN & AIDA \\
\hline
Distance headway ($d$) & Yes & Yes & Yes & 
Yes \\
Relative speed ($\Delta v$) & Yes & Yes & Yes & Yes \\
Speed ($v$) & Yes & No & No & No \\
$\tau^{-1}$ & No & Yes & Yes & Yes \\
\hline
\end{tabular}
\end{table}

\begin{table}[!htb]
\caption{Model parameter counts}\label{tab:param_count}
\centering
\begin{tabular}{lcccc}
\hline
& IDM & BC-MLP & BC-RNN & AIDA \\
\hline 
Count & 6 & 4125 & 6465 & 7670 \\
\hline
\end{tabular}
\end{table}

\begin{table}[!htb]
\caption{Fitted IDM parameters: mean and standard deviations across 15 seeds.}\label{eq:hyperparams}
\centering
\begin{tabular}{ccc}
\hline
$\Tilde{v}$ & $\tau$ & $d_{0}$ \\
\hline
12.2 $\pm$ 0.2 & 0.83 $\pm$ 0.03 & 1.07 $\pm$ 0.07 \\
\hline
$a_{max}$ & $b$ & $\sigma$ \\
\hline
0.21 $\pm$ 0.006 & 2.68 $\pm$ 0.19 & 0.46 $\pm$ 0.004 \\
\hline
\end{tabular}
\end{table}

\begin{table}[h!]
\caption{Two-sided Welch's t-test results of offline MAE-IQM against baseline models. Asterisks indicate statistical significance with $\alpha=0.05$.}
\label{tab:offline_eval}
\centering
\begin{tabular}{llcc}
\hline
Baseline & Comparison & t(df=14) & p-value \\ \hline
IDM      & dense-lane  & t=16.38 & p\textless{}0.001* \\
BC-MLP   & dense-lane  & t=29.74 & p\textless{}0.001* \\
BC-RNN   & dense-lane  & t=16.03 & p\textless{}0.001* \\
IDM      & sparse-lane   & t=29.11 & p\textless{}0.001* \\
BC-MLP   & sparse-lane   & t=0.44 & p=0.66             \\
BC-RNN   & sparse-lane   & t=-0.04 & p=0.97             \\ \hline
\end{tabular}
\end{table}

\begin{table}[h!]
\caption{Two-sided Welch's t-test results of online ADE-IQM against baseline models. Asterisks indicate statistical significance with $\alpha=0.05$.}
\label{tab:online_eval}
\centering
\begin{tabular}{lllc}
\hline
Baseline & Comparison & t(df=14) & p-value \\ \hline
IDM      & dense-lane  & t=3.05 & p\textless{}0.01* \\
BC-MLP   & dense-lane  & t=-5.46 & p\textless{}0.001* \\
BC-RNN   & dense-lane  & t=8.73 & p\textless{}0.001* \\
IDM      & sparse-lane   & t=58.18 & p\textless{}0.001* \\
BC-MLP   & sparse-lane   & t=-3.77 & p\textless{}0.001* \\
BC-RNN   & sparse-lane   & t = 6.87 & p\textless{}0.001* \\ \hline
\end{tabular}
\end{table}

\subsection{Simulation platform}
We built a single-agent simulator for online evaluation of the trained models. The simulator plays back lead vehicle trajectories from the dataset which is converted into the Frenet frame to compute LV related observations for the ego vehicle. We model the effect of ego control actions on its own position and velocity in the Frenet frame using linear dynamics:
\begin{align}
    \begin{bmatrix}
         x' \\
         y' \\
         v_{x}' \\
         v_{y}' \\
    \end{bmatrix} = 
    \begin{bmatrix}
        1 & 0 &\Delta t & 0 \\
        0 & 1 & 0 & \Delta t \\
        0 & 0 & 1 & 0 \\
        0 & 0 & 0 & 1 \\
    \end{bmatrix}
    \begin{bmatrix}
         x \\
         y \\
         v_{x} \\
         v_{y} \\
    \end{bmatrix} + 
    \begin{bmatrix}
        0.5\Delta t^2 & 0 \\
        0 & 0.5\Delta t^2 \\
        \Delta t & 0 \\
        0 & \Delta t \\
    \end{bmatrix}
    \begin{bmatrix}
    a_{x} \\
    a_{y}
    \end{bmatrix}
\end{align}

Since the models only control longitudinal action $a_{x}$, we used a simple feedback controller for lateral actions with position and velocity gain $[-0.01, -0.2]$.

\begin{figure}[!htb]
    \centering
\includegraphics[width=0.4\textwidth]{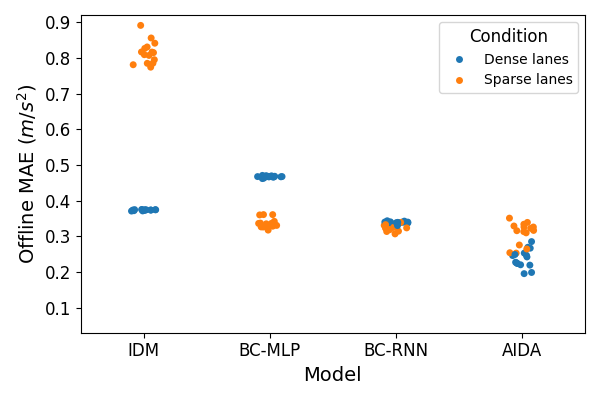}
\includegraphics[width=0.4\textwidth]{figs_rebuttal/offline_mae_iqm.png}
    \caption{Comparison of offline MAE by each model with IQM (top) and without IQM (bottom).}  \label{fig:offline_ade_iqm_comparison}
\end{figure}

\begin{figure}[!htb]
    \centering
\includegraphics[width=0.4\textwidth]{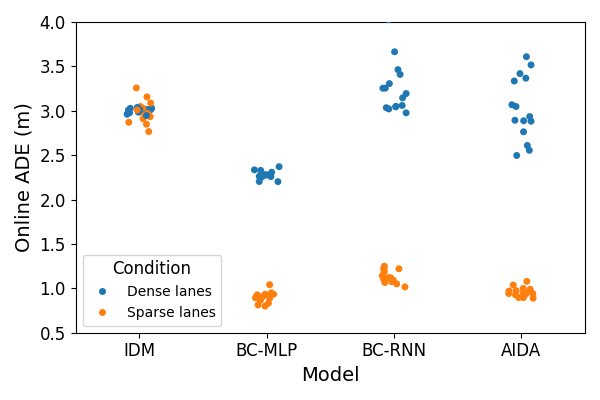}
\includegraphics[width=0.4\textwidth]{figs_rebuttal/online_mae_iqm.png}
    \caption{Comparison of online ADE of each model with IQM (top) and without IQM (bottom).}   \label{fig:online_ade_iqm_comparison}
\end{figure}

\subsection{Comparison of average metrics and IQM}\label{sec:iqm_comparison}
This section compares model offline (Figure \ref{fig:offline_ade_iqm_comparison}) and online (Figure \ref{fig:online_ade_iqm_comparison}) evaluation performance with and without IQM. In the offline evaluation setting, IQM did not substantially affect model performance, e.g., the IDM MAE in the sparse-lane setting was only reduced by 0.1 $m/s^2$. In the online evaluation setting, IQM reduced the average ADE of BC-RNN and AIDA by 1 m and substantially reduced the upper tail of AIDA seeds. However, IDM also increased the difference between IDM's ADE in the dense-lane and sparse-lane settings, which is consistent with IDM having worse offline prediction accuracy in the sparse-lane setting as shown in Figure \ref{fig:offline_mae}. Overall, IQM did not change the ranking of models.
It should be noted that in the estimated IDM model, the parameter estimate $a_{max}$  is significantly lower than that reported in \cite{Hoogendorn_2010}.

\subsection{Statistical testing results}\label{sec:statistical_tests}
Statistical tests using the two-sided Welch's t-test with 5 percent rejection level are shown in Table \ref{tab:offline_eval} and \ref{tab:online_eval} for offline and online evaluations, respectively. 

\end{document}